\documentclass[acmsmall, screen]{acmart}
\usepackage{amsmath,amssymb,amsfonts}
\usepackage{algorithmic}
\usepackage{graphicx}
\usepackage{xspace}
\usepackage{tabularx}
\usepackage{array}
\usepackage{booktabs}
\usepackage{algorithm}
\newcommand{\midsepremove}{\aboverulesep=0mm \belowrulesep=0mm}
\midsepremove
\newcommand{\midsepdefault}{\aboverulesep=0.6mm \belowrulesep=0.9mm}
\midsepdefault
\usepackage{xcolor}
\usepackage{colortbl}
\usepackage{multirow}

\newcommand*{\eg}{\textit{e.g.}\@\xspace}
\newcommand*{\ie}{\textit{i.e.}\@\xspace}
\makeatletter
\newcommand*{\etc}{%
    \@ifnextchar{.}%
        {etc}%
        {etc.\@\xspace}%
}
\newcommand*{\etal}{%
    \@ifnextchar{.}%
        {et al}%
        {et al.\@\xspace}%
}
\makeatother

\newboolean{showcomments}
\setboolean{showcomments}{true} 
\ifthenelse{\boolean{showcomments}}
{ \newcommand{\mynote}[3]{
    \fbox{\bfseries\sffamily\scriptsize#1}
    {\small$\blacktriangleright$\textsf{\emph{\color{#3}{#2}}}$\blacktriangleleft$}}}
{ \newcommand{\mynote}[3]{}}
\newcommand{\shrink}[1]{}

\graphicspath{{./figs/}}

\newcommand{\topone} {{top-1}\xspace}
\newcommand{\topfive} {{top-5}\xspace}
\newcommand{\bleu} {BLEU\xspace}
\newcommand{\bleups} {BLEU-PS\xspace}
\newcommand{\rouge} {Rouge\xspace}

\newcommand{\DNN} {\texttt{DNN}\xspace}
\newcommand{\DNNs} {\texttt{DNNs}\xspace}
\newcommand{\CNN} {\texttt{CNN}\xspace}

\newcommand{\NMT} {\texttt{NMT}\xspace}

\newcommand{\CNNs} {\texttt{CNNs}\xspace}
\newcommand{\NN} {\texttt{KNN}\xspace}

\newcommand{\SVM} {\texttt{SVM}\xspace}
\newcommand{\DT} {\texttt{DT}\xspace}
\newcommand{\NB} {\texttt{NB}\xspace}
\newcommand{\oracle} {\texttt{Oracle}\xspace}

\newcommand{\mn}[1]{\texttt{#1}}
\newcommand{\premodel} {\texttt{premodel}\xspace}
\newcommand{\ML} {\texttt{ML}\xspace}
\newcommand{\dnninput} {\textit{input}\xspace}
\newcommand{\dnninputs} {\textit{inputs}\xspace}

\newcommand{\cparagraph}[1]{\vspace{1mm}\noindent \textbf{#1}}

\definecolor{Gray}{gray}{0.98}

\setcopyright{acmcopyright}
\acmDOI{10.475/123_4}
\acmISBN{123-4567-24-567/08/06}

\emergencystretch=\maxdimen
\begin{document}

\title{Optimizing Deep Learning Inference on Embedded Systems Through Adaptive Model Selection}

\author{Vicent Sanz Marco}
\authornote{Both co-authors contributed equally to this research.}
\affiliation{%
  \institution{Osaka University}
   \country{Japan}
  }
\email{v.sanzmarco@cmc.osaka-u.ac.jp}

\author{Ben Taylor}
\authornotemark[1]
\affiliation{%
  \institution{Lancaster University}
  \country{United Kingdom}
  }
\email{b.d.taylor@lancaster.ac.uk}


\author{Zheng Wang}
\affiliation{%
  \institution{University of Leeds}
   \country{United Kingdom}
  }
\email{z.wang5@leeds.ac.uk}

\author{Yehia Elkhatib}
\affiliation{%
  \institution{Lancaster University}
   \country{United Kingdom}
  }
\email{y.elkhatib@lancaster.ac.uk}

%


\thanks{
A preliminary version of this article appeared in ACM LCTES 2018~\cite{taylor2018adaptive}. Accepted to be published at ACM TECS.}


\begin{abstract}
Deep neural networks (\DNNs) are becoming a key enabling technique for many application domains. However, on-device inference on
battery-powered, resource-constrained embedding systems is often infeasible due to prohibitively long inferencing time and resource
requirements of many \DNNs. Offloading computation into the cloud is often unacceptable due to privacy concerns, high latency, or the
lack of connectivity. While compression algorithms often succeed in reducing inferencing times, they come at the cost of reduced
 accuracy.

This paper presents a new, alternative approach to enable efficient execution of \DNNs on embedded devices. Our approach dynamically
determines which \DNN to use for a given input, by considering the desired accuracy and inference time. It employs machine learning to
develop a low-cost predictive model to quickly select a pre-trained \DNN to use for a given input and the optimization constraint. We
achieve this by first off-line training a predictive model, and then using the learned model to select a \DNN model to use for new, unseen
inputs.
We apply our approach to two representative \DNN domains: image classification and machine translation. We evaluate our approach on a
Jetson TX2 embedded deep learning platform, and consider a range of influential \DNN models including convolutional and recurrent neural
networks. For image classification, we achieve a 1.8x reduction in inference time with a 7.52\% improvement in accuracy, over the
most-capable single \DNN model. For machine translation, we achieve a 1.34x reduction in inference time over the most-capable single
model, with little impact on the quality of translation.

\end{abstract}



\setcopyright{acmlicensed} \acmJournal{TECS} \acmYear{2019} \acmVolume{1} \acmNumber{1} \acmArticle{1} \acmMonth{1}
\acmPrice{15.00}\acmDOI{10.1145/3371154}

\maketitle

\section{Introduction}
Deep learning is getting a lot of attention recently, and with good reason. It has proven ability in solving many difficult problems such
as object recognition~\cite{donahue14,he2016deep}, facial recognition~\cite{parkhi2015deep,sun2014deep}, speech
processing~\cite{pmlrv48amodei16}, and machine translation~\cite{bahdanau2014neural}. While many of these tasks are also important
application domains for embedded systems~\cite{lane2018deep}, existing deep learning solutions are often resource intensive tasks,
consuming a considerable amount of CPU, GPU, memory, and power~\cite{CanzianiPC16}. Without a solution, the hoped-for advances on smart
embedded sensing will not be realized.

Numerous optimization tactics have been proposed to enable deep inference\footnote{Inference in this work means
applying a pre-trained model on an input to obtain the corresponding output.} on embedded devices. Prior approaches are
either architecture specific~\cite{song2017towards}, or come with drawbacks. Model compression is a commonly used
technique for accelerating deep neural networks (\DNNs). Using compression, a \DNN can be optimized by reducing its
resource and computational
requirements~\cite{han2015learning,han2016eie,howard2017mobilenets,Georgiev:2017:LMA:3139486.3131895}. Unfortunately,
this also comes at the cost of a reduction in model accuracy. To avoid this, alternate approaches have been developed;
offload some, or all, computation to a cloud server where resources are available for fast inference
times~\cite{Kang2017neurosurgeon,teerapittayanon2017distributed}. This, however, is not always possible due to high
network latency or poor reliability~\cite{elkhatib2015building}.
Furthermore, sending sensitive data over a network could be prohibited due to privacy constraints.

Our work seeks to offer an alternative to execute pre-trained \DNN models on embedded systems. Our aim is to design a \emph{generalizable}
approach to optimize \DNNs to run \emph{efficient} inference without affecting accuracy. Central to our approach is an adaptive scheme for
determining, \emph{at runtime}, which of the available \DNNs is the best fit for the input and evaluation criterion. Our key insight is
that the optimal model -- the model which is able to give the correct input in the fastest time -- depends on the input data and the
evaluation criterion. In fact, by utilizing multiple \DNN models we are able to increase accuracy in some cases. In essence, for a simple
input -- an image taken under good lighting conditions, with a contrasting background; or a short sentence with little punctuation -- a
simple, fast model would be sufficient; a more complex input would require a more complex model. Similarly, if an accurate output with high
confidence is required, a more sophisticated but slower model would have to be employed -- otherwise, a simple model would be good enough.

In this work, we employ machine learning (\ML) to \emph{automatically} construct a predictor able to dynamically select the optimum model
to use. Our predictor is first trained \emph{off-line}. Then, using a set of automatically tuned features of the \DNN input, the predictor
determines the optimum \DNN for a \emph{new}, \emph{unseen} input; taking into consideration the evaluation criterion. We show that our
approach can automatically derive high-quality heuristics for different evaluation criteria. The learned strategy can effectively leverage
the prediction capability and runtime overhead of candidate \DNNs, leading to an overall better accuracy when compared with the most
capable \DNN model, but with significantly less runtime overhead. Compression can be used in conjunction to our approach to generate
multiple \DNNs of varying capability, then automatically choose the best at runtime. This is a new way for optimizing deep inference on
embedded devices.

Our approach is designed to be generally applicable to all domains of deep learning. As case studies, we choose two typical and unique
domains for evaluation: image classification and machine translation. Both domains have a dynamic range of available \DNN architectures
including convolutional and recurrent neural networks. We evaluate our approach on the NVIDIA Jetson TX2 embedded platform and consider a
wide range of influential \DNN models,  ranging from simple to complex. Experimental results show that our approach delivers portable good
performance across the two \DNN tasks. For image classification, it improves the inference accuracy by 7.52\% over the most-capable single
\DNN model while achieving 1.8x less inference time. For machine translation, it reduces the inference time of 1.34x over the most-capable
model with negligible impact on the quality of the translation.

%
%
\newpage

The paper makes the following technical contributions:
\begin{itemize}
\item We present a novel \ML based approach to automatically learn how to select \DNN models based on the input and precision requirement
    (Section~\ref{sec:approach}). Our system has little training overhead  as it does not require any modification to pre-trained \DNN
    models;

\item Our work is the first to leverage multiple \DNN models to improve the prediction accuracy and reduce inference time on embedded systems (Section~\ref{sec:results}).

\item Our approach has a good generalization ability as it works effectively on different network architectures, application domains and
    input datasets. We show that our approach can be easily integrated with existing model compression techniques to improve the overall
    results.

\end{itemize}

\section{Motivation}

As a motivation, consider two contrasting examples, image classification and machine translation, of using \DNNs. The experiments in this
section are carried out on a NVIDIA Jetson TX2 platform where we use the GPU for inference; full details of the system can be seen in
Section~\ref{sec:eval_setup}.

\begin{figure*}[t!]
\def\arraystretch{0.8}
    \centering
    \begin{tabularx}{1\textwidth} {>{\centering\arraybackslash}m{1.0in}>{\centering\arraybackslash}m{1.0in}>{\centering\arraybackslash}m{1.0in}>{\centering\arraybackslash}m{1.9in}}

        \includegraphics[width=0.19\textwidth]{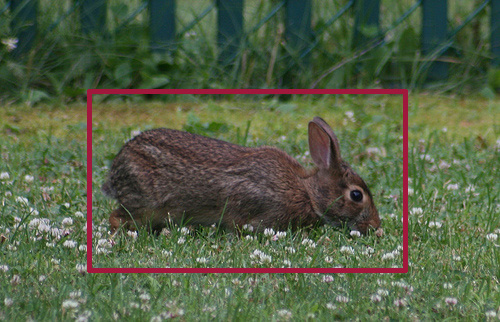} &
        \includegraphics[width=0.19\textwidth]{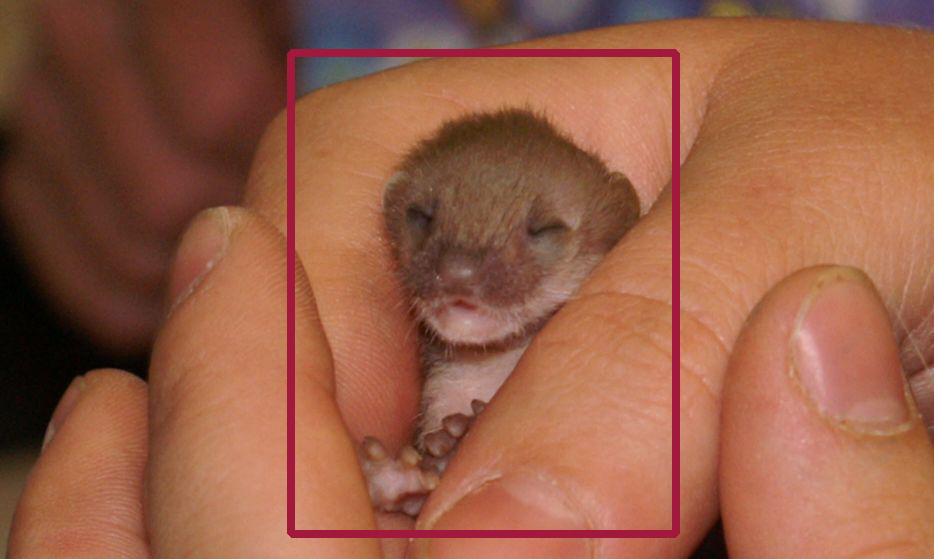} &
        \includegraphics[width=0.19\textwidth]{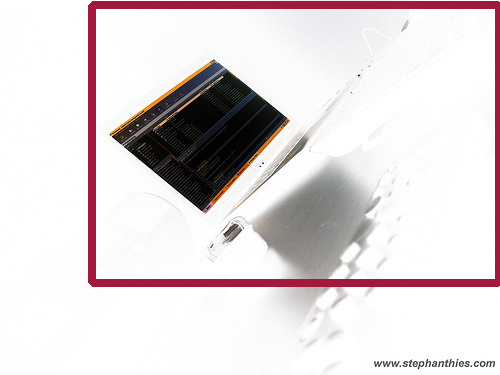} &
        \includegraphics[width=0.29\textwidth]{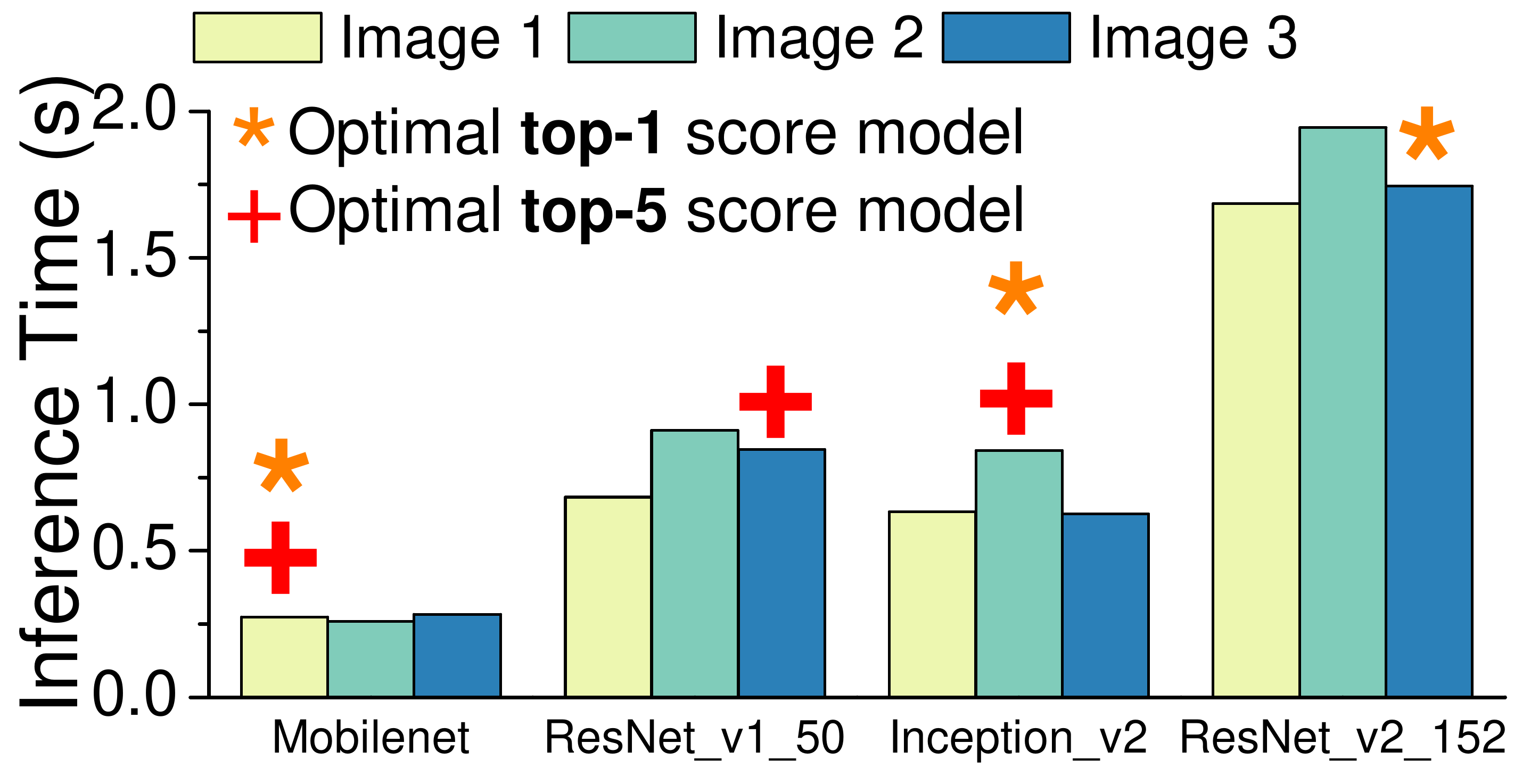} \\
        \vspace{-2mm}
        {\centering \scriptsize (a) Image 1} &
        {\centering \scriptsize (b) Image 2} &
        {\centering \scriptsize (c) Image 3} &
        {\centering \scriptsize (d) Inference time}
    \end{tabularx}
    \caption{The inference time (d) of four CNN-based image recognition models when processing images (a) - (c).
    The target object is highlighted on each image.
    This example (combined with Table~\ref{tbl:motivation_model_accuracy}) shows that the optimal model (\ie the fastest one that gives an accurate output)
    depends on the success criterion and the input.}
    \label{fig:motivation_example}
\end{figure*}

\begin{table}[t!]
	\scriptsize
	\centering
	\caption{List of models that give the correct prediction per image under the \topfive and
    the \topone scores.}
	\midsepremove
	\begin{tabularx}{0.95\textwidth}{p{2cm}*{4}{X}}
		\toprule
            & Image 1 & Image 2 & Image 3\\
            \midrule

            \rowcolor{Gray} \topfive score & \mn{\textbf{MobileNet\_v1\_025}}, \mn{ResNet\_v1\_50}, \mn{Inception\_v2}, \mn{ResNet\_v2\_152} & \mn{\textbf{Inception\_v2}}, \mn{ResNet\_v1\_50},
            \mn{ResNet\_v2\_152} & \mn{\textbf{ResNet\_v1\_50}}, \mn{ResNet\_v2\_152}\\

            \topone score & \mn{\textbf{MobileNet\_v1\_025}}, \mn{ResNet\_v1\_50}, \mn{Inception\_v2}, \mn{ResNet\_v2\_152} & \mn{\textbf{Inception\_v2}},
            \mn{ResNet\_v2\_152} & \mn{\textbf{ResNet\_v2\_152}}\\
            \bottomrule
	\end{tabularx}
	\label{tbl:motivation_model_accuracy}
\end{table}
\subsection{Image Classification}  \label{sec:motivation}


\vspace{-2mm} \cparagraph{Setup.} For image classification, we investigate one subset of \DNNs: Convolutional Neural Networks (\CNNs). We
compare the performance of three influential \CNN architectures: \mn{Inception}~\cite{ioffe2015batch}, \mn{ResNet}~\cite{he2016identity},
and \mn{MobileNet}~\cite{howard2017mobilenets}\footnote{ Each model architecture follows its own naming convention.
\mn{MobileNet\_v$i$\_$j$},
 where $i$ is the version number, and $j$ is a width multiplier out of 100, with 100 being the full uncompressed model.
\mn{ResNet\_v$i$\_$j$}, where $i$ is the version number, and $j$ is the number of layers in the model.
\mn{Inception\_v$i$}, where $i$ is the version number.}.
Specifically, we
used the following models: \mn{MobileNet\_v1\_025},
the \mn{MobileNet} architecture with a width multiplier of 0.25;
\mn{ResNet\_v1\_50},
the first version of \mn{ResNet} with 50 layers;
\mn{Inception\_v2},
the second version of \mn{Inception};
and \mn{ResNet\_v2\_152},
the second version of \mn{ResNet} with 152 layers.
All these models are built upon TensorFlow~\cite{tensorflow} and  have been pre-trained by
independent researchers using the ImageNet ILSVRC 2012 \emph{training dataset}~\cite{ILSVRC15}.

\cparagraph{Evaluation criteria.} Each model takes an image as input and returns a list of label confidence values as output. Each value
indicates the confidence that a particular object is in the image. The resulting list of object values are sorted in descending order
regarding their prediction confidence, so that the label with the highest confidence appears at the top of the list. In this example, the
accuracy of a model is evaluated using the \topone and the \topfive scores defined by the ImageNet Challenge. Specifically, for the \topone
score, we check if the top output label matches the ground truth label of the primary object; and for the \topfive score, we check if the
ground truth label of the primary object is in the top 5 of the output labels for each given model.

\cparagraph{Results.} Figure~\ref{fig:motivation_example}d shows the inference time per model using three images from the ImageNet ILSVRC
\emph{validation dataset}. Recognizing the main object (a cottontail rabbit) from the image shown in Figure~\ref{fig:motivation_example}a
is a straightforward task. We can see from Table~\ref{tbl:motivation_model_accuracy} that all models give the correct answer under the
\topfive and \topone score criterion. For this image, \mn{MobileNet\_v1\_025} is the best model to use under the \topfive score, because it
has the fastest inference time -- 6.13x faster than \mn{ResNet\_v2\_152}.
Clearly, for this image, \mn{MobileNet\_v1\_025} is good enough,
and there is no need to use a more advanced (and expensive) model for inference.
If we consider a slightly more complex object
recognition task shown in Figure~\ref{fig:motivation_example}b, we can see that \mn{MobileNet\_v1\_025} is unable to give a correct answer
regardless of our success criterion. In this case \mn{Inception\_v2} should be used, although this is 3.24x slower than
\mn{MobileNet\_v1\_025}. Finally, consider the image shown in Figure~\ref{fig:motivation_example}c, intuitively it can be seen that
this is a more difficult image recognition task, as the main object is a similar color to the background. In this case the optimal model
changes depending on our success criterion. \mn{ResNet\_v1\_50} is the best model to use under \topfive scoring, completing
inference 2.06x faster than \mn{ResNet\_v2\_152}. However, if we use \topone for scoring we must use \mn{ResNet\_v2\_152}  to obtain the
correct label, despite it being the most expensive model. Inference time for this image is 2.98x and 6.14x slower than
\mn{MobileNet\_v1\_025} for \topfive and \topone scoring respectively.
The results are similar if we use different images of similar complexity levels.

\begin{figure*}[t!]
\def\arraystretch{0.8}
    \centering
    \begin{tabularx}{1\textwidth} {>{\centering\arraybackslash}m{0.46\textwidth}>{\centering\arraybackslash}m{0.46\textwidth}}

        \includegraphics[width=0.45\textwidth]{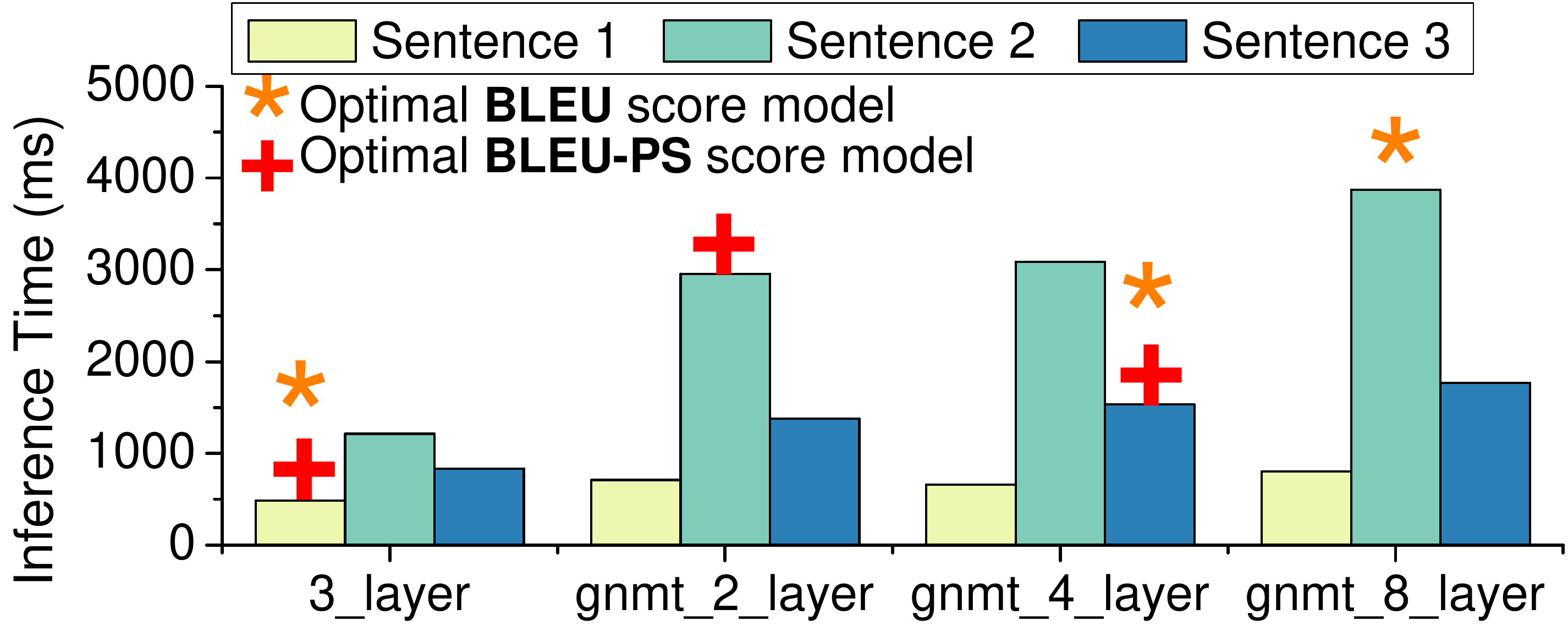} &
        \includegraphics[width=0.45\textwidth]{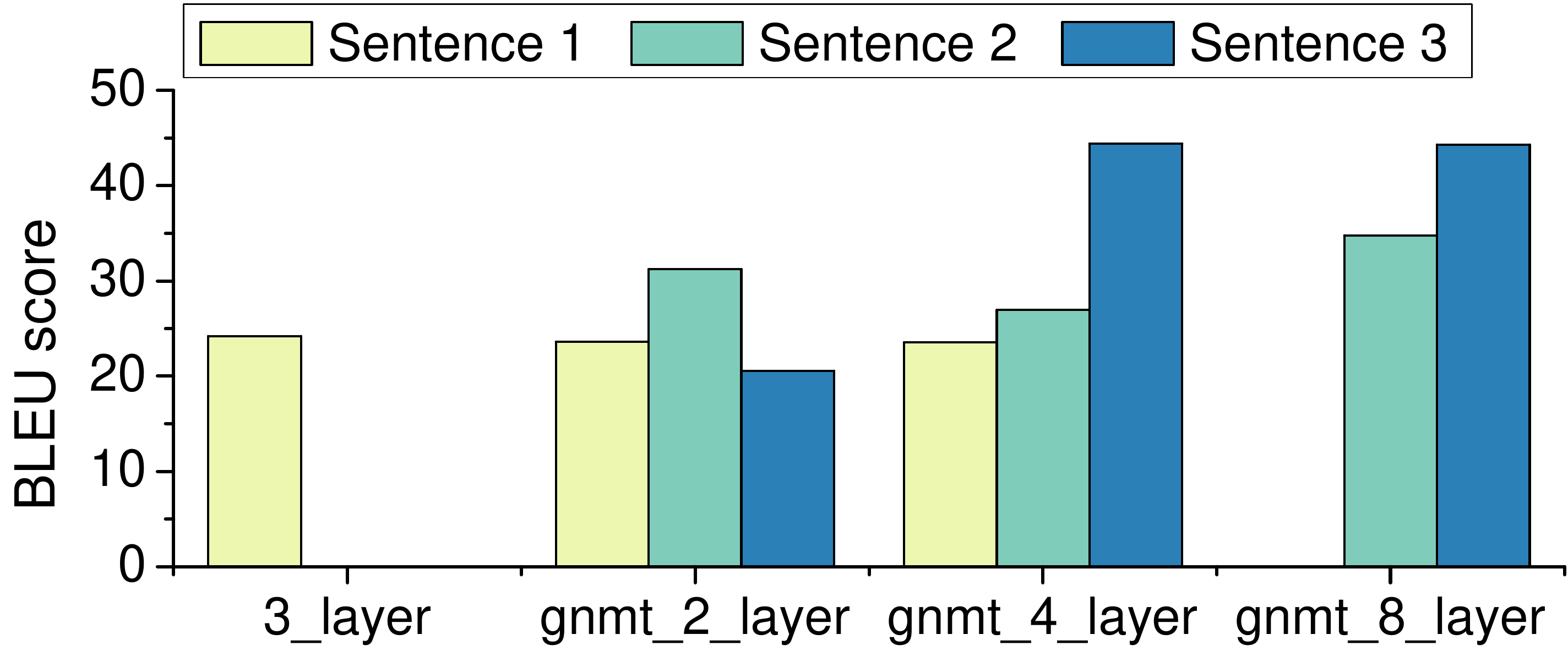} \\
        {\centering \scriptsize (a) Runtime} &
        {\centering \scriptsize (b) \bleu scores} \\
    \end{tabularx}
    \caption{The inference time, optimal model (a), and \bleu score (b) of three sentences (shown in Table~\ref{tbl:motivation_sentences}). Here the optimal model achieves the highest score for an evaluation criteria.
    Model names explained in footnote~\ref{note:MT_models}.}
    \label{fig:MT_motivation}
\end{figure*}

\vspace{-2mm}
\subsection{Machine Translation}

\vspace{-2mm} \cparagraph{Setup.} In the second experiment, we consider the following 4 machine translation models as they provide a range
of accuracy and runtime capabilities \footnote{\label{note:MT_models} We name our models using the following convention:
\texttt{\{gnmt\_\}N\_layer}, we prefix the name with \texttt{gnmt\_}where the model uses the Google Neural Machine Translation
Attention~\cite{45610}, and $N$ is the number of layers in the model.}: \texttt{3\_layer}, \texttt{gnmt\_2\_layer},
\texttt{gnmt\_4\_layer}, and \texttt{gnmt\_8\_layer}. We chose three distinct sentences from the WMT15/16 English-German newstest
dataset~\cite{wmt}, which can be seen in Table~\ref{tbl:motivation_sentences}.

\cparagraph{Evaluation criteria.} Unlike image classification, no metrics similar to \topone and \topfive exist for machine translation.
Therefore, we use the following metrics for evaluation:

\begin{itemize}
\item \emph{\textbf{\bleu} (higher is better)}. Bilingual Evaluation Understudy is widely used to evaluate machine translation model
    output. It returns a value between 0 and 1, 1 being a perfect output; it is very rarely achieved.

\item \emph{\textbf{\bleups} (higher is better)}. \bleu per second. \bleu is only able to represent a degree of correctness, we also use \bleups to evaluate the trade-off between \bleu and inference time.
\bleups is similar to Energy Delay Product (EDP, which is used to evaluate the trade-off between energy consumption and runtime), and is calculated as $\frac{BLEU \times BLEU} {Infer. Time}$.

\end{itemize}

\vspace{-2mm}
\cparagraph{Results.}
Figure~\ref{fig:MT_motivation} shows the inference time, \bleu score and optimal model for each sentence.
\textit{Sentence 1}, is the simplest sentence, therefore the easiest translation task.
The optimal model for all metrics is our simplest, \texttt{3\_layer}.
Surprisingly, our most complex model, \mn{gnmt\_8\_layer}, fails on this sentence; by using the cheapest model we achieve a higher accuracy 1.66x quicker.
Similarly the optimal model for \textit{Sentence 3} across both metrics is \mn{gnmt\_4\_layer}.
In this case, we cannot use our cheapest model, as it fails.
By choosing the optimal model for \textit{Sentence 3} we can infer 1.15x quicker, without impacting accuracy.
It is clear that \textit{Sentence 2} is more complex than \textit{Sentence 1}, it is much longer, has frequent punctuation, and contains non-words, \eg 2008 and TV.
In this case, the optimal model changes depending on the evaluation metric.
If we are optimising for \bleups we use \mn{gnmt\_2\_layer}, which is 1.31x times quicker than \mn{gnmt\_8\_layer}.
However, if we would like to maximize accuracy, we need to use \mn{gnmt\_8\_layer}.

\begin{table*}[t!]
	\def\arraystretch{1.05}
	\scriptsize
	\centering
	\vspace{-2mm}
	\caption{The sentences used in Figure~\ref{fig:MT_motivation}}
	\vspace{-4mm}
	\midsepremove
	\begin{tabularx}{\textwidth}{lX}
		\toprule
		\textbf{Sentence ID}    & \textbf{Sentence}                                             \\
		\midrule
		\rowcolor{Gray}     \textit{1}              & High on the agenda are plans for greater nuclear co-operation.\\
		\textit{2}              & Advertisements, documentaries, TV series and parts in films consumed his next decade but after his 2008 BBC series, LennyHenry.tv, he thought: " What are you going to do next, Len, because it all feels a bit like you're marking time or you're slightly going sideways."\\
		\rowcolor{Gray}     \textit{3}              & Kenya has started biometrically registering all civil servants in an attempt to remove "ghost workers" from the government's payroll.\\
		\bottomrule
	\end{tabularx}
	\label{tbl:motivation_sentences}
	\vspace{-3mm}
\end{table*}

\vspace{-2mm}
\subsection{Summary of Motivation Experiments}
The above examples show that the best model depends on the input and the evaluation criterion.
Hence, determining which model to use is non-trivial.
What we need is a technique that can automatically choose the most efficient model to use for any
given input.
In the next section, we describe our adaptive approach that solves this task.





\section{Our Approach \label{sec:approach}}

\begin{figure}[t!]
\centering
\begin{minipage}{.38\textwidth}
  \centering
  \includegraphics[width=\textwidth]{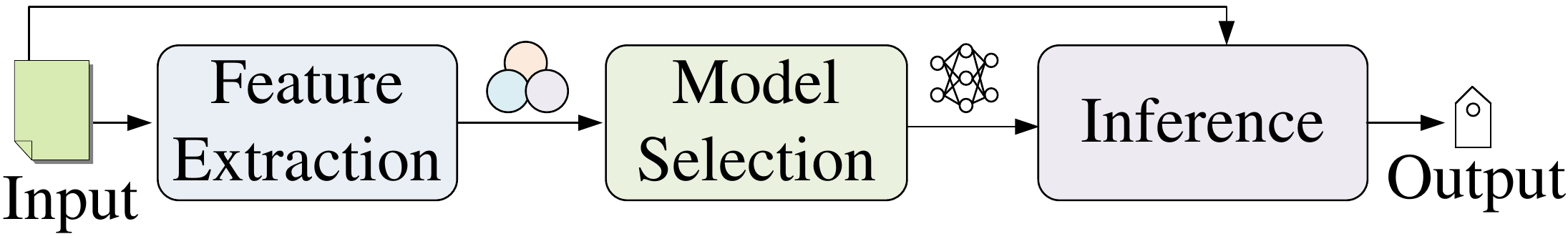}
  \caption{Overview of our approach.} \label{fig:overview}
\end{minipage}%
\hfill
\begin{minipage}{.6\textwidth}
  \centering
  \includegraphics[width=\textwidth]{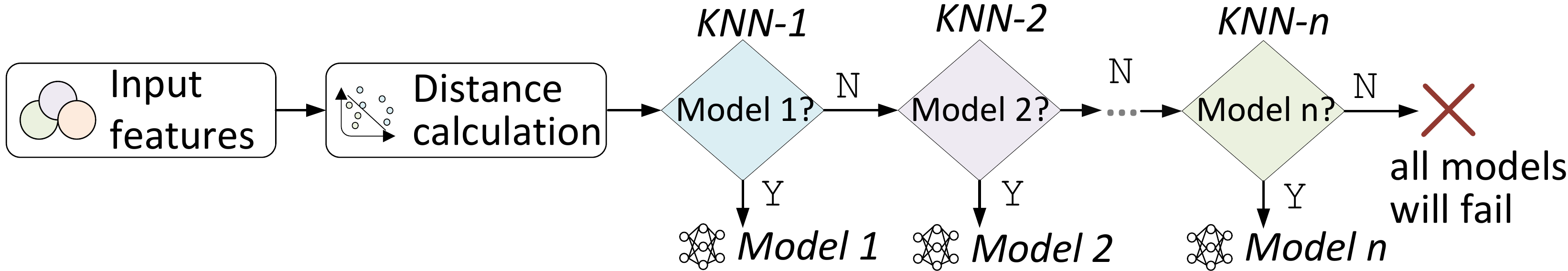}
  \caption{Our \premodel for image classification, made up of a series of \NN models predicting whether to use a specific \DNN or not. Our process for selecting classifiers is described in Section~\ref{sec:classifier_selection}.}
  \label{fig:IC_premodel_design}
\end{minipage}
\end{figure}

\subsection{Overview}
Figure~\ref{fig:overview} depicts the overall workflow of our approach. Our approach trades memory footprints for accuracy and reduced
inference time. At the core of our approach is a predictive model (termed \premodel) that takes a \emph{new, unseen} \dnninput (\eg an
image or sentence), and predicts which of a set of pre-trained \DNN models to use for that given \dnninput. This decision may vary
depending on the scoring method used at the time, \eg either \topone or \topfive in image classification.

Our \premodel is automatically generated based on the problem domain. An example of a generated \premodel can be seen in
Figure~\ref{fig:IC_premodel_design}. The prediction of our \premodel is based on a set of quantifiable properties -- or \emph{features},
such as the number of edges and brightness of an image -- of the \dnninput. Once a model is chosen, the \dnninput is passed to the selected
\DNN, which then performs inference on the \dnninput. Finally, the inference output of the selected \DNN is returned. Use of our \premodel
will work in exactly the same way as any single model \ie the input and output will be in the same format, however, we are able to
dynamically select the best model to use.

%

\vspace{-3mm}
\subsection{Premodel Design} \label{sec:model_desc}

To design an effective \premodel for embedded inference, we consider two design goals:  (i) high accuracy and (ii) fast execution time. By
correctly choosing the optimal model, a highly accurate \premodel can reduce the average inference time. Furthermore, a fast \premodel is
important because if a \premodel takes much longer than any single \DNN will be useless. The task of choosing a candidate \DNN to use is
essentially a classification problem in machine learning. Although using a standard \ML classifier as a \premodel can yield acceptable
results, we discovered we can maximize performance by changing the \premodel architecture depending on the domain (see
Section~\ref{sec:classifier_selection}).

In this work we consider four well-established classifiers: K-Nearest Neighbour (\NN), a simple clustering based classifier; Decision Tree
(\DT), a tree based classifier; Naive Bayes (\NB), a probabilistic classifier; and Support Vector Machine (\SVM), a more complex, but well
performing classification algorithm. In Section~\ref{sec:alt_premodels}, we evaluate a number of different \ML techniques, including
Decision Trees, Support Vector Machines, and \CNNs.

Simultaneously, we consider two different types of \premodel architecture: (i) A simple, single classifier architecture using only one \ML
classifier to predict which \DNN to use; (ii) A multiple classifier architecture (See Figure~\ref{fig:IC_premodel_design}), a sequence of
\ML classifiers where each classifier predicts whether to use a single \DNN or not. The later is described in more detail in
Section~\ref{sec:multi_architecture}. Finally, we chose a set of features to represent each \dnninput; the selection process is detailed in
in Section~\ref{sec:features}.

\vspace{-1mm}
\subsubsection{Multiple Classifier Architecture} \label{sec:multi_architecture}

Figure~\ref{fig:IC_premodel_design} gives an overview of a \premodel implementing a multiple classifier architecture. As an example, we
will use the \NN based \premodel created for image classification.  For each \DNN model we wish to include in our \premodel, we use a
separate \NN model. As our \NN models are going to contain much of the same data, we begin our \premodel by calculating our K closest
neighbours. Taking note of which record of training data each of the neighbours corresponds to, we avoid recalculating the distance
measurements; instead, we simply change the labels of these data-points. \textit{KNN-1} is the first \NN model in our \premodel, through
which all input to the \premodel will pass. \textit{KNN-1} predicts whether the input image should use \textit{Model-1} to classify it or
not. If \textit{KNN-1} predicts that \textit{Model-1} should be used, then the \premodel returns this label, otherwise the features are
passed on to the next \NN, \ie \textit{KNN-2}. This process repeats until the image reaches \textit{KNN-n}, the final \NN model in our
\premodel. In the event that \textit{KNN-n} predicts that we should not use \textit{Model-n}, the next step will depend on the user's
declared preference: (i) use a pre-specified model, to receive some output to work with; or (ii) do not perform inference and inform the
user of the failure.



\subsection{Inference Model Selection} \label{sec:classifier_selection}
\begin{algorithm}[t!]
	\caption{Model Selection}
	\label{alg:classifier_selection}
    \scriptsize
	\begin{algorithmic}[1]
	\REQUIRE \textit{data, $\theta$, selection\_method}
	\STATE $Model\_1\_DNN = most\_optimum\_DNN(data)$
	\STATE $curr\_DNNs.add(Model\_1\_DNN)$
	\STATE $curr\_acc = get\_acc(curr\_DNNs)$
	\STATE \textit{acc\_diff = 100}
	\WHILE{$\textit{acc\_diff} > \theta$}
		\STATE \textit{improvement\_metric = next\_selection\_metric(selection\_method)}
		\STATE \textit{next\_DNN = greatest\_improvement\_DNN(data, curr\_DNNs, improvement\_metric)}
		\STATE $curr\_DNNs.add(next\_DNN)$
		\STATE $new\_acc = get\_acc(curr\_DNNs)$
		\STATE \textit{acc\_diff = new\_acc - curr\_acc}
		\STATE $curr\_acc = new\_acc$
	\ENDWHILE
	\end{algorithmic}
\end{algorithm}

\vspace{-2mm}
In Algorithm~\ref{alg:classifier_selection} we describe our selection process for choosing which \DNNs to
include in our \premodel. This algorithm takes in three parameters: (1) \textit{data}, containing the output of each
\DNN for every \dnninput; (2) $\theta$, a threshold parameter telling us when to terminate the model selection process;
and (3) \textit{selection\_method}, one of a choice of methods that produces an \textit{improvement\_metric} (accuracy
or optimal) for determining when if a candiate \DNN should be included in the \premodel in each iteration. We consider
the following three model selection methods:

\begin{itemize}
\item \emph{\textbf{Based on accuracy.}}  Using this selection method, we will add a \DNN to \premodel if it has the greatest improvement
    in accuracy for each iteration. There are some cases where the selected \DNNs all fail to make a correct prediction, but some of the
    remaining candidate models can. During each selection iteration, we will choose a remaining \DNN that if it is included, it can lead
    to the most significant improvement in prediction accuracy for \premodel.

\item \emph{\textbf{Based on optimal.}} In each iteration of the loop, the most optimal \DNN is selected; \ie the one that gives the
    greatest overall improvement in accuracy, but leads to the lowest increase in inference time, for the selected \DNN set.

\item \emph{\textbf{Alternate.}} A hybrid of the first two approaches. We alternate between choosing the most optimal and the most accurate
\DNN in each iteration. Our first \DNN is always the most optimal.
\end{itemize}

\vspace{-1mm} \cparagraph{Model selection process.} The model selection process works as follows.
\vspace{-1mm}
\begin{itemize}

\item \emph{\textbf{Initialization.}} The first \DNN we include is the most optimal model for our training data, i.e., the \DNN that
    is most frequently considered to be optimal across training instances.

\item \emph{\textbf{Iterative selection.}} At each iteration, we consider each of the remaining potential \DNNs, and add the one which
    brings the greatest improvement to our \textit{improvement\_metric} (accuracy or optimal), which can change per iteration based on
    the \textit{selection\_method}.

\item \emph{\textbf{Termination.}} We iteratively add new \DNNs until our accuracy improvement is lower than the termination threshold,
    $\theta$\%.

\end{itemize}
Using this Model Selection Algorithm we are able to add \DNNs that best compliment one another when working together, maximizing our
accuracy while keeping
    our runtime low. In Section~\ref{sec:msa_eval} we evaluate the impact of different parameter choices on our algorithm.

\begin{figure}
	\centering
	\begin{minipage}{.47\textwidth}
		\centering
			\includegraphics[width=1\textwidth]{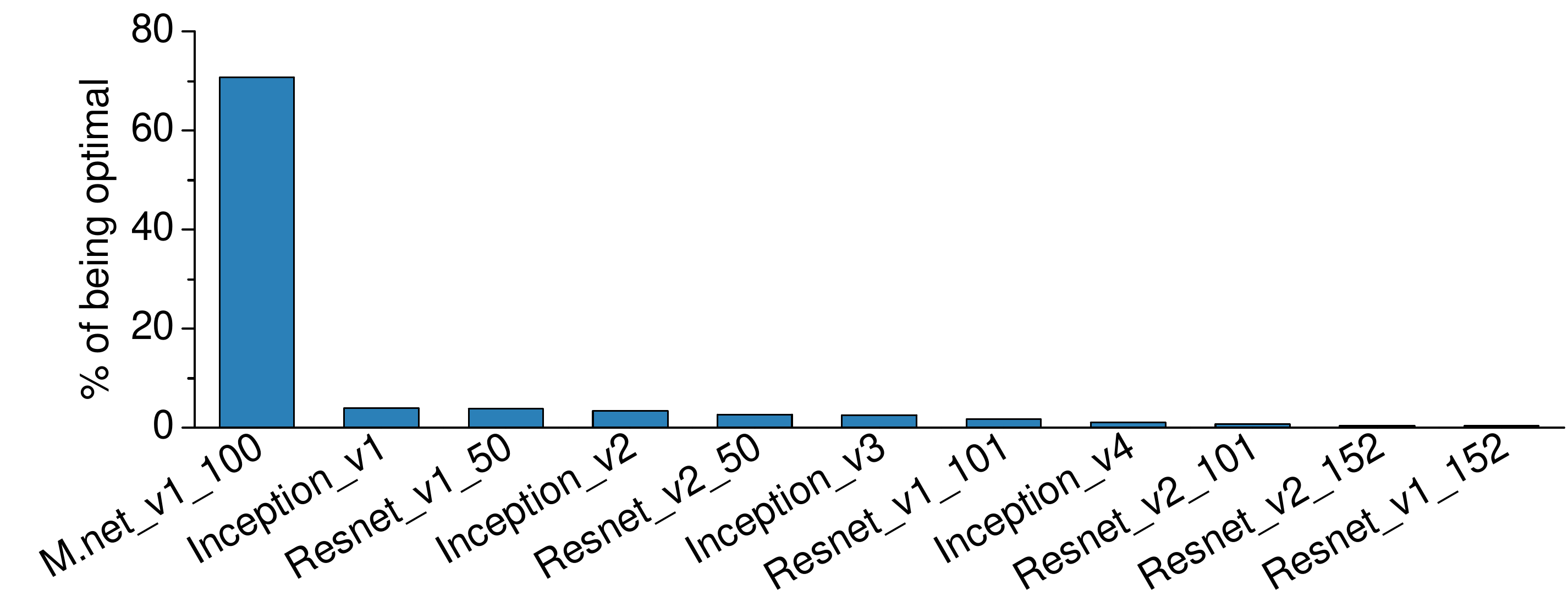}\\
		\caption{How often a \CNN model is considered to be optimal under \topone on the training dataset.}
		\label{fig:optimum_models}
	\end{minipage}%
	\hfill
	\begin{minipage}{.47\textwidth}
		\centering
		\includegraphics[width=1\textwidth]{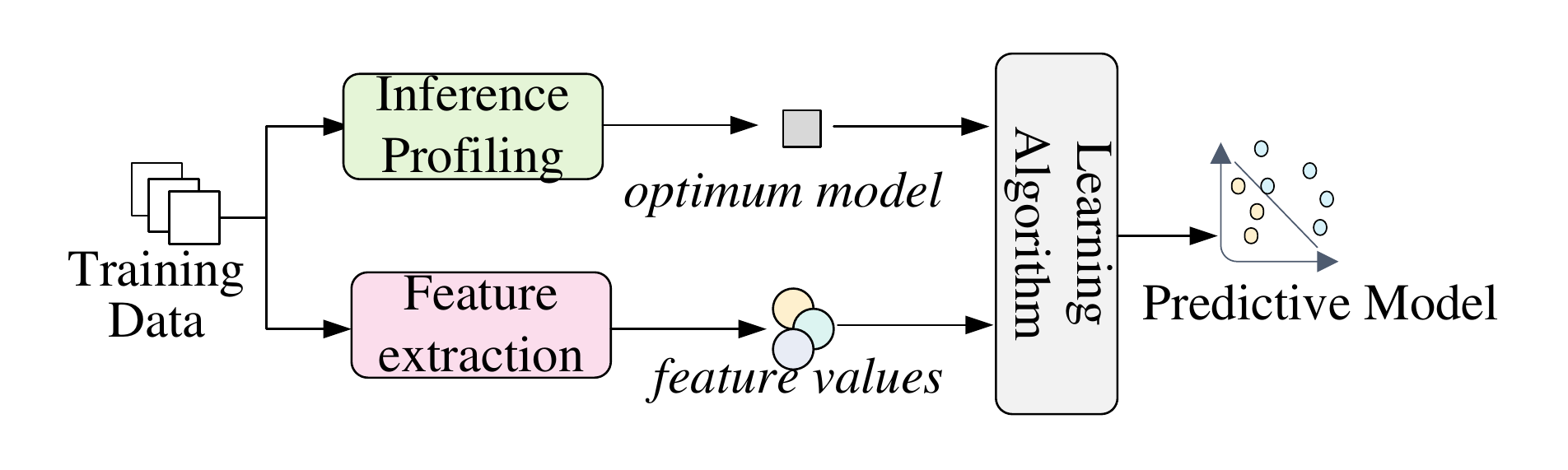}\\
		\caption{The training process. We use the same procedure to train each individual model within the \premodel for each evaluation criterion.}
		\label{fig:training}
	\end{minipage}
\end{figure}

\begin{figure*}[t!]
\def\arraystretch{0.82}
	\centering
	\begin{tabularx}{1\textwidth} {cXX}
		\includegraphics[width=0.32\textwidth]{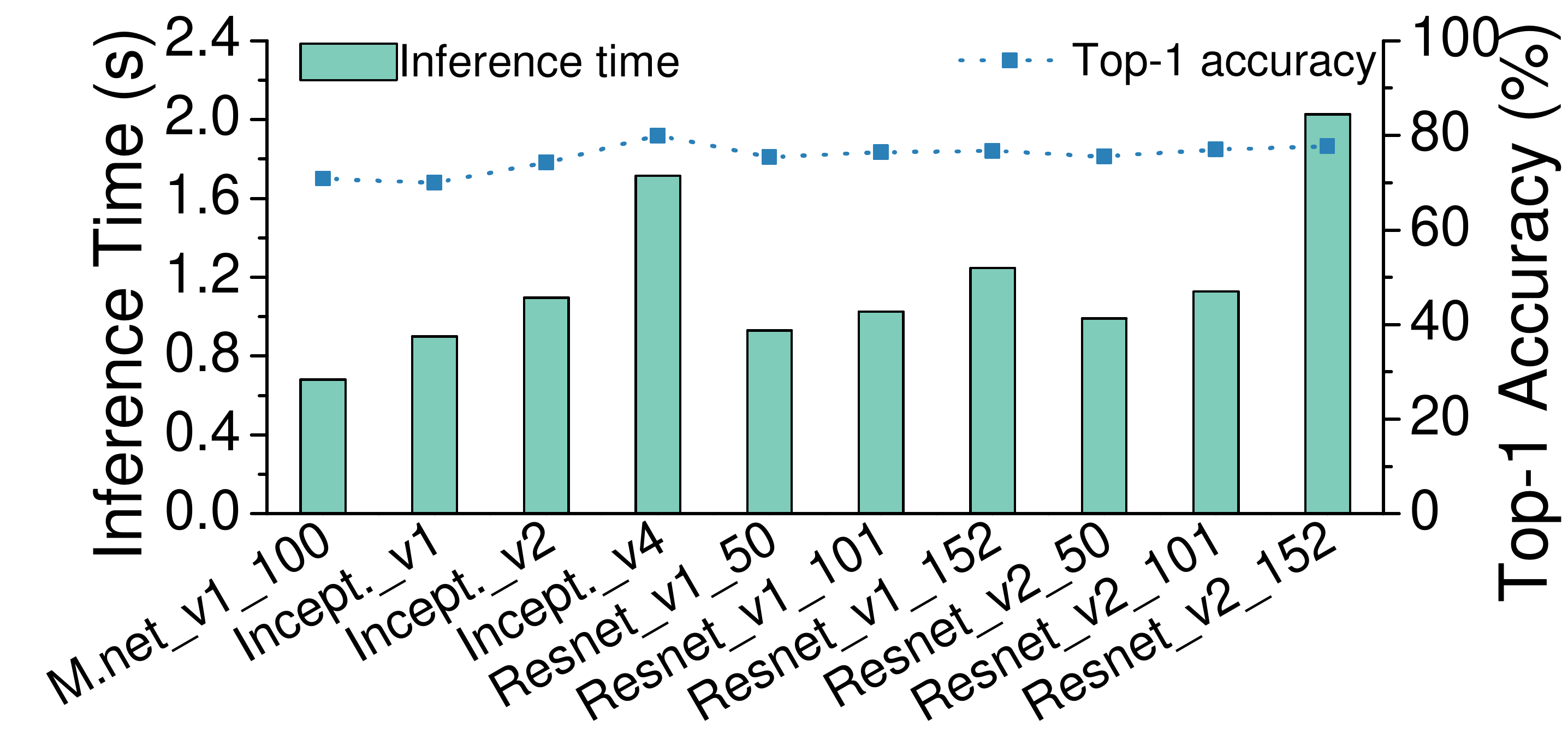} &
		\includegraphics[width=0.31\textwidth]{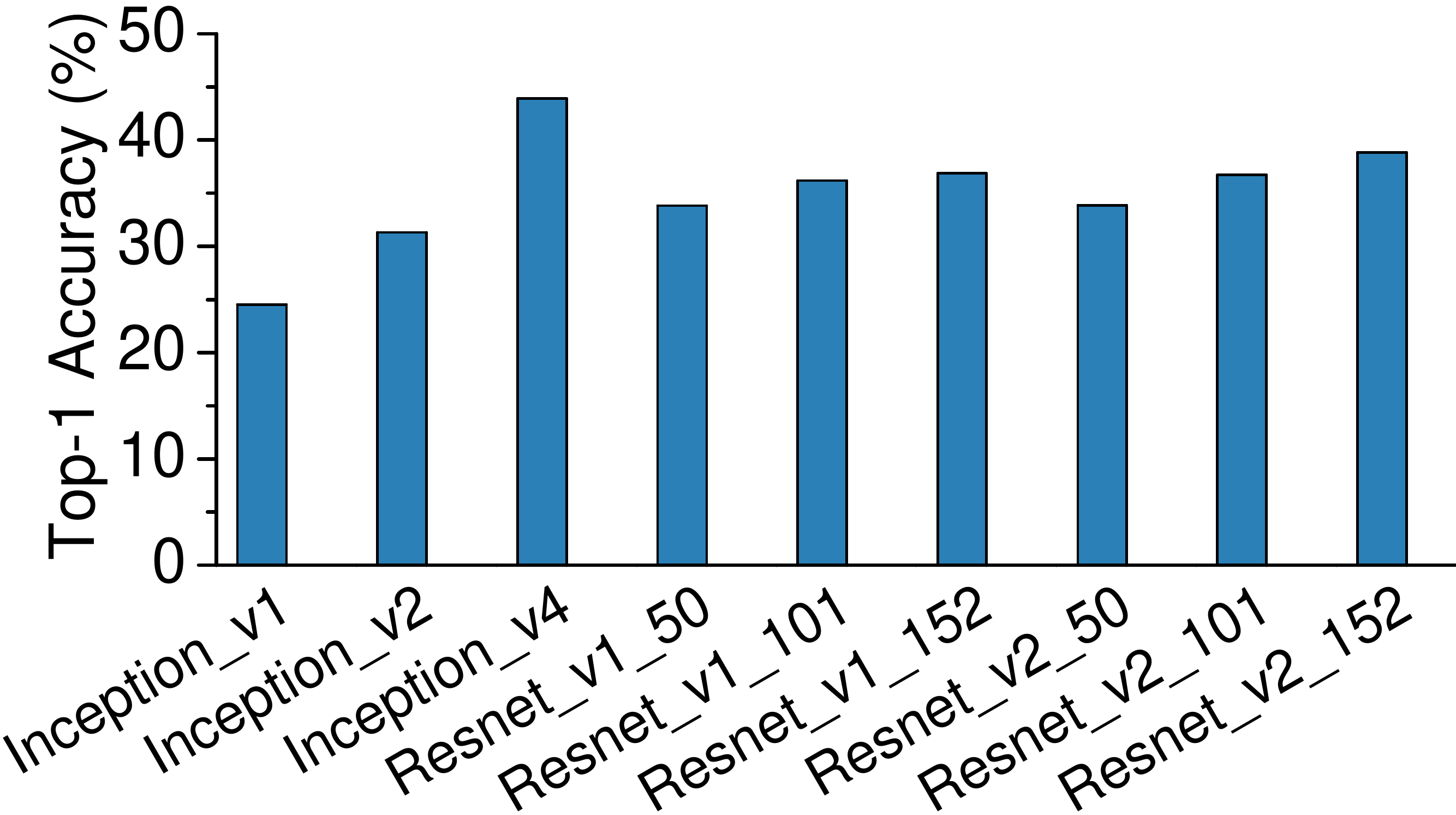}&
		\includegraphics[width=0.31\textwidth,clip]{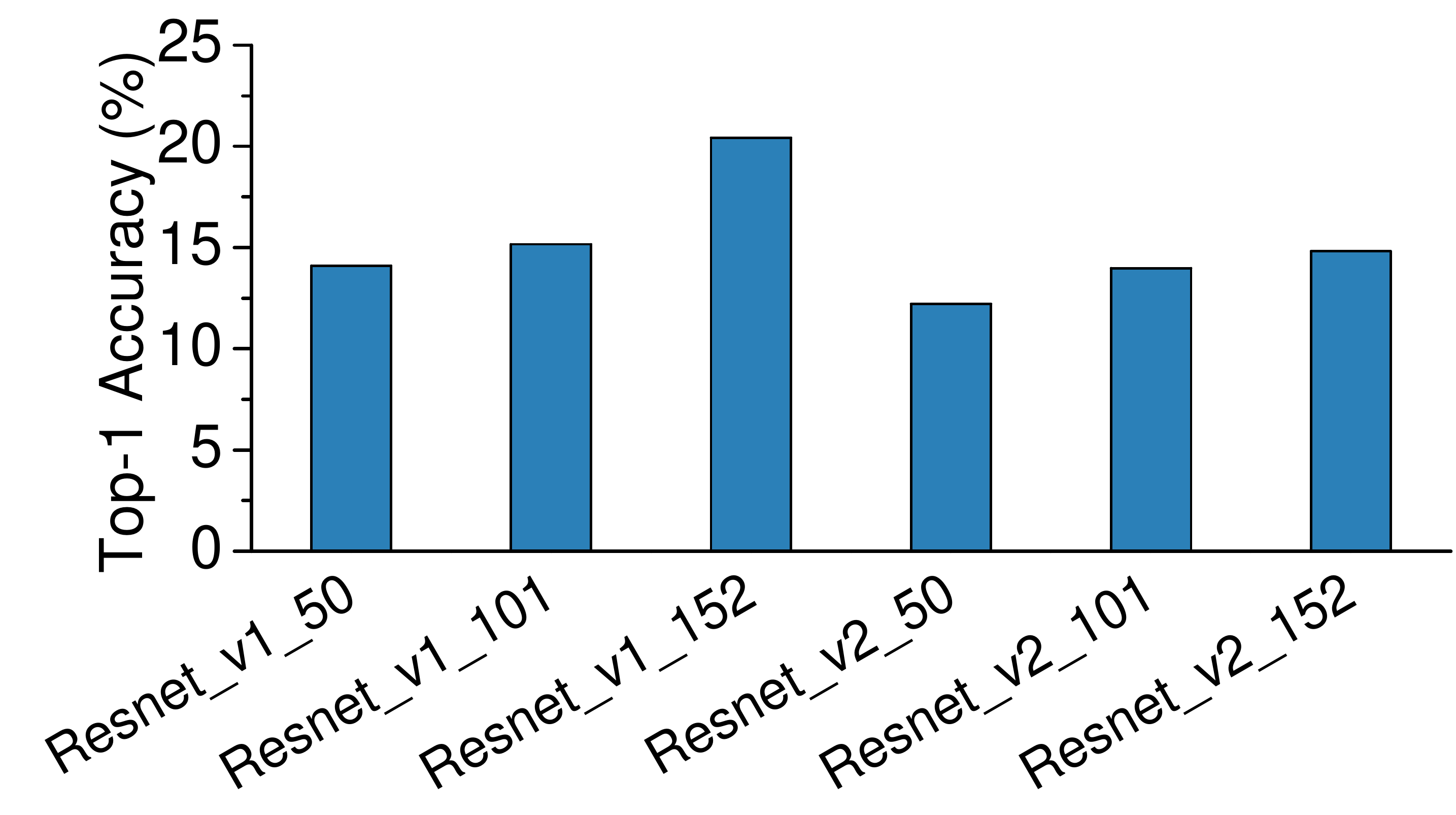}\\
		{\centering \scriptsize (a) All \CNNs} &
		{\centering \scriptsize (b) Where \texttt{MobileNet} fails} &
		{\centering \scriptsize (c) Where \texttt{Mobilnet} \& \texttt{Inception} fails.} \\

	\end{tabularx}
	\caption{Image classification results. (a) The \topone accuracy and inference time of all \CNNs we consider.
	(b) The \topone accuracy of all \CNNs on the images on which \mn{MobileNet\_v1\_100} fails.
	(c) The \topone accuracy of all \CNNs on the images on which \mn{MobileNet\_v1\_100} and \mn{Inception\_v4} fails.}
	\label{fig:classifier_selection}
\end{figure*}

\cparagraph{Illustrative example.} We now walk through the Model Selection Algorithm using the image classification problem as an example.
In this example, we set our threshold $\theta$ to 0.5, which is empirically decided through our pilot experiments. We also set
\textit{selection\_method} to ``based on accuracy" for this example. \emph{We carry out a a sensitivity analysis for these parameters later
in Section~\ref{sec:msa_eval}.} Figure~\ref{fig:optimum_models} shows the percentage of our training data that considers each of our \CNNs
to be optimal. For this example, the model selection process works as follows:

\begin{itemize}
\vspace{-1mm}
\item \emph{\textbf{First model.}} The first model is the most optimal model. In this example, \mn{MobileNet\_v1\_100} is chosen to be
    \textit{Model-1} because it is optimal for most (70.75\%) of our training data.

\item \emph{\textbf{Iterative selection.}} If we were to follow the ``based on optimal" selection method and choose the next most optimal
    \CNN, we would choose \mn{Inception\_v1}. However, we do not do this as it would result in our \premodel being formulated of many
    cheap, yet inaccurate models. Instead we choose to look at the training data and consider which of our remaining \CNNs gives the
    greatest improvement in accuracy (i.e., ``based on accuracy"), as '\textit{Accuracy}' is our \textit{improvement\_metric}.
    Intuitively, as image classification is either right or wrong, we are searching for the \CNN that is able to correctly classify the
    most of the remaining 29.25\% cases where \mn{MobileNet\_v1\_100} fails. As seen in Figure~\ref{fig:classifier_selection}b,
    \mn{Inception\_v4} is best, correctly classifying 43.91\% of the remaining data and creating a 12.84\% increase in \premodel
    accuracy. Repeating this process (Figure~\ref{fig:classifier_selection}c), we add \mn{ResNet\_v1\_152} to our \premodel, further
    increasing total accuracy by 2.55\%.

\item \emph{\textbf{Termination.}} After adding \mn{ResNet\_v1\_152}, we iterate once more to achieve a \premodel accuracy increase of
    less than 0.5\% ($\theta$), and therefore terminate.

\item \emph{\textbf{Results.}} After running this process, our \premodel is composed of: \mn{MobileNet\_v1\_100} for \textit{Model-1},
    \mn{Inception\_v4} for \textit{Model-2}, and \mn{ResNet\_v1\_152} for \textit{Model-3}.

\end{itemize}


\vspace{-4mm}
\subsection{Training the Premodel} \label{sec:premodel_training}
Training our \premodel follows the standard procedure, and is a multi-step process. We describe the entire training process in detail
below, and provide a summary in Figure~\ref{fig:training}. Generally, we need to figure out which candidate \DNN is optimum for each of our
training \dnninputs (to be used by the Model Selection Algorithm described in Section~\ref{sec:classifier_selection}), we then train our
\premodel to predict the same for any \emph{new}, \emph{unseen} \dnninputs.

\cparagraph{Generate training data.} Our training dataset consists of the feature values and the corresponding optimum \DNN for each
\dnninput under an evaluation criterion. To evaluate the performance of the candidate \DNN models, they must be applied to unseen
\dnninputs.
We exhaustively execute each candidate \DNN on the \dnninputs, measuring the inference time and prediction results.
Inference time is measured on an unloaded machine to reduce noise; it is a one-off cost -- \ie it only needs to be completed once.
Because the relative runtime of models is stable, training can
be performed on a high-performance server to speedup data generation.
It is to note that adding a new \DNN simply requires executing all \dnninputs on the new \DNN while taking the same measurements described above.

Using the execution time, and \DNN output results, we can calculate the \textit{optimum} classifier for each \dnninput; \ie
the model that achieves the accuracy goal (\topone, \topfive, or BLEU) in the least amount of time.
Finally, we extract the feature values
(described in Section~\ref{sec:features}) from each \dnninput, and pair the feature values to the optimum classifier for each \dnninput,
resulting in our complete training dataset.

\cparagraph{Building the premodel.} The training data is used to determine the classification models to use and their optimal
hyper-parameters. All classifiers we consider for \premodel support are supervized learning algorithms. Therefore, we simply supply the
classifier with the training data and it carries out its internal algorithm. For example, in \NN classification the training data is used
to give a label to each point in the model, then during prediction the model will use a distance measure (in our case we use Euclidian
distance) to find the K nearest points (in our case K=5). The label with the highest number of points to the prediction point is the output
label.

\cparagraph{Training cost.} Total training time of our \premodel is dominated by generating the  training data, which took less than a day
using a NVIDIA P40 GPU on a multi-core server. This can vary depending on the number of candidate inference models to be included. In our
case, we had an unusually long training time as we considered a large number of \DNN models. We would expect in deployment that the user
has a much smaller search space for potential \DNNs. The time in model selection and parameter tuning is negligible (less than 2 hours) in
comparison. See also Section~\ref{sec:overhead}.


\vspace{-1mm}
\subsection{Features} \label{sec:features}
One key aspect in building a successful predictor is selecting the right features to characterize the input. In this work, we have
developed an automatic feature selection process, the user is simply required to provide a number of candidate features. Automatic feature
generation could be used to provide candidate features, however this is out of the scope of this work.



\subsubsection{Feature Selection} \label{sec:feature_sel}
Feature extraction is the biggest overhead of our \premodel, therefore by reducing our feature count we can decrease the total execution
time. Moreover, by reducing the number of features we are also improving the generalizability of our \premodel.

Initially, we use correlation-based feature selection.
If pairwise correlation is high for any pair of features, we drop one of them and keep the other; retaining most of the information.
We perform this by constructing a matrix of correlation coefficients using Pearson product-moment correlation (\textit{PCC}).
The coefficient value falls between $-1$ and $+1$.
The closer the absolute value is to $1$, the stronger the correlation between the two features being tested.
We set a threshold of 0.75 and removed any features that had an absolute \textit{PCC} higher than the threshold.

Next, we evaluated the importance of each of our remaining features. To do so, we first trained and evaluated our \premodel using K-Fold
cross validation (see also Section~\ref{sec:overhead}) and all of our current features, recording \premodel accuracy. We then remove each
feature and re-evaluate the model on the remaining features, taking note of the change in accuracy. If there is a large drop in accuracy
then the feature must be important, otherwise, the feature does not hold much importance for our purposes. Using this information we
performed a greedy search, removing the least important features one by one. We detail the outcome of this process in
Section~\ref{sec:feature_importance}. Below we have summarized the result of each feature selection stage on both of our case studies.
Removing any of the remaining features resulted in a significant drop in model accuracy.



\subsubsection{Feature Scaling} The final step before passing our features to a \ML model is scaling each feature to a common range (between 0 and 1) in order to prevent the range of any single feature being a factor in its importance.
Scaling does not affect the distribution or variance of feature values.
To achieve this during deployment, we record the minimum and maximum values of each feature in the training dataset and use these to scale the corresponding features of new data.




\subsection{Runtime Deployment}
Deployment of our proposed method is designed to be simple and easy to use, similar to current \DNN usage techniques.
We have encapsulated all of the inner workings, such as needing to read the output of the \premodel and then
choosing the correct \DNN model.
A user would interact with our \premodel by simply calling a prediction function and getting a result in return in the same format as the \DNNs in use.
Using image classification as an example, the return value would be the predicted labels and their confidence levels.

\section{Evaluation Setup}
We apply our approach to two representative \DNN domains: image classification and machine translation. Each domain is presented as a case
study (Sections~\ref{sec:case1} and \ref{sec:case2}) that shows the results at each stage of applying our approach; providing an
end-to-end analysis. The case studies will end with an analysis in Section~\ref{sec:results} of how our approach performed against other
representative \DNNs in the domain. In the remainder of this section, we describe our evaluation setup and methodology.

\vspace{-2mm}
\subsection{Hardware and Software \label{sec:eval_setup}}
\vspace{-1mm}
\cparagraph{Hardware.} We evaluate our approach on the NVIDIA Jetson TX2 embedded deep learning platform. The system has a 64~bit dual-core
Denver2 and a 64~bit quad-core ARM Cortex-A57 running at 2.0~Ghz, and a 256-core NVIDIA Pascal GPU running at 1.3~Ghz. The board has 8~GB
of LPDDR4 RAM and 96~GB of storage (32~GB eMMC plus 64~GB SD card).

\vspace{-1mm} \cparagraph{System software.} Our evaluation platform runs Ubuntu 16.04 with Linux kernel v4.4.15. We use Tensorflow v.1.0.1,
cuDNN (v6.0) and CUDA (v8.0.64). Our \premodel is implemented using the Python scikit-learn package. Our feature extractor is built upon
OpenCV and SimpleCV.



\vspace{-2mm}
\subsection{Evaluation Methodology \label{sec:method}}

\subsubsection{Model Evaluation} We use \emph{10-fold cross-validation} to evaluate each \premodel on its respective dataset. Specifically,
we split our dataset into 10 sets which equally represent the full dataset, \eg if we consider image classification, we partition the 50K
validation images into 10 equal sets, each containing 5K images. We retain one set for testing our \premodel, and the remaining 9 sets are
used as training data. We repeat this process 10 times (folds), with each of the 10 sets used exactly once as the testing data. This
standard methodology evaluates the generalization ability of a machine-learning model.

We evaluate our approach using the following metrics:

\vspace{-1.5mm}
\begin{itemize}
\item \emph{\textbf{Inference time} (lower is better)}. Wall clock time between a model taking in an input and producing an output,
    including the overhead of our \premodel.

\item \emph{\textbf{Energy consumption} (lower is better)}. The energy used by a model for inference. For our approach, this also
    includes the energy consumption of the \premodel. We deduct the static power when the system is idle.

\item \emph{\textbf{Accuracy} (higher is better)}. The ratio of correctly labeled cases to the total number of testing cases.
\end{itemize}

\vspace{-1.5mm} \cparagraph{Metrics for image classification.} The following metrics are specific to image classification:
\begin{itemize}
\item \emph{\textbf{Precision} (higher is better)}. The ratio of a correctly predicted images to the total number of images that are predicted to have a
    specific object. This metric answers e.g., ``\emph{Of all the images that are labeled to have a cat, how many actually have a cat?}".

\item \emph{\textbf{Recall} (higher is better)}. The ratio of correctly predicted images to the total number of test images that belong to an object class.
    This metric answers e.g., ``\emph{Of all the test images that have a cat, how many are actually labeled to have a cat?}".

\item \emph{\textbf{F1 score} (higher is better)}.  The weighted average of Precision and Recall, calculated as $2\times\frac{Recall
    \times Precision} {Recall + Precision}$. It is useful when the test datasets have an uneven distribution of classes.
\end{itemize}

\vspace{-1.5mm} \cparagraph{Metrics for machine translation.} The following metrics are specific to machine translation:
\begin{itemize}
\item \emph{\textbf{BLEU} (higher is better)}.  Similar to precision in image classification.
It is a measure of how much the words (and/or n-grams) in the \DNN model output appeared in the reference output(s).

\item \emph{\textbf{Rouge} (higher is better)}. Similar to recall in image classification.
It is a measure of how much the words (and/or n-grams) in the reference output(s) appear in the \DNN model output.

\item \emph{\textbf{F1 measure} (higher is better)}. Similar to F1 score for image classification.
The weighted average of BLEU and Rouge, calculated as $2\times\frac{Rouge
    \times BLEU} {Rouge + BLEU}$.
\end{itemize}

\subsubsection{Performance Report} We report the \emph{geometric mean} of the aforementioned evaluation metrics across the cross-validation
folds. To collect inference time and energy consumption, we run each model on each input repeatedly until the 95\% confidence bound per
model per input is smaller than 5\%. In the experiments, we exclude the loading time of the \DNN models as they only need to be loaded once
in practice. However, we include the overhead of our \premodel in all our experimental data. To measure energy consumption, we developed a
lightweight runtime to take readings from the onboard energy sensors at a frequency of 1,000 samples per second.
It is to note that our work does not directly optimize for energy consumption. We found that in our scenario there is  little difference
when optimizing for energy consumption compared to time.


\vspace{-1mm}
\section{Case Study 1: Image Classification}
\label{sec:case1} To evaluate our approach in the domain of image classification we consider 14 pre-trained \CNN models from the
TensorFlow-Slim library~\cite{silberman2013tensorflow}. The models are built using TensorFlow and trained on the ImageNet ILSVRC 2012
\emph{training set}. We use the Imagenet ILSVRC 2012 \emph{validation set} to create the training data for our \premodel, and evaluate it
using \emph{cross-validation} (see Section~\ref{sec:method}).

\begin{table}[t!]
	\parbox{.45\linewidth}{
		\small
		\def\arraystretch{0.82}
		\scriptsize
		\centering
		\captionsetup{justification=centering}
		\caption{All features considered for image classification.}
		\vspace{-3mm}
		\midsepremove
		\begin{tabular}{ll}
			\toprule
			\textbf{Feature}                    & \textbf{Description}                                    \\
			\midrule
			\rowcolor{Gray}     \textit{n\_keypoints}               & \# of keypoints                                         \\
			\textit{avg\_brightness}            & Average brightness                                      \\
			\rowcolor{Gray}     \textit{brightness\_rms}            & Root mean square of brightness                          \\
			\textit{avg\_perc\_brightness} & Average of perceived brightness                         \\
			\rowcolor{Gray}     \textit{perc\_brightness\_rms} & Root mean square of perceived  brightness               \\
			\textit{contrast}                   & The level of contrast                                   \\
			\rowcolor{Gray}     \textit{edge\_length\{1-7\}}        & A 7-bin histogram of edge lengths                       \\
			\textit{edge\_angle\{1-7\}}         & A 7-bin histogram of edge angles                        \\
			\rowcolor{Gray}     \textit{area\_by\_perim}            & Area / perimeter of the main object                     \\
			\textit{aspect\_ratio}              & The aspect ratio of the main object                     \\
			\rowcolor{Gray}     \textit{hue\{1-7\}}                 & A 7-bin histogram of the different hues               \\
			\bottomrule
		\end{tabular}
		\label{tbl:all_features}
	}
	\hfill
	\parbox{.45\linewidth}{
		\small
		\centering
		\def\arraystretch{0.82}
		\scriptsize
		\captionsetup{justification=centering}
		\caption{Correlation values (absolute) of removed features to the kept ones for image classification.}
		\vspace{-3mm}
		\midsepremove
		\begin{tabular}{lll}
			\toprule
			\textbf{Kept Feature}												& \textbf{Removed Feature}            			&\textbf{Correl.}		\\
			\midrule
			\rowcolor{Gray}													& \textit{perc\_brightness\_rms}				& 0.98 					\\
			\rowcolor{Gray}													& \textit{avg\_brightness}						& 0.91 					\\
			\rowcolor{Gray} 	\multirow{-3}{*}{\textit{avg\_perc\_brightness}}	& \textit{brightness\_rms}						& 0.88		 			\\
			\textit{edge\_length1}				&      \textit{edge\_length \{4-7\}}						& 0.78 - 0.85 					\\						
			\rowcolor{Gray} 	\emph{hue1}												& \textit{hue \{2-6\}}								& 0.99 					\\
			\midrule
			\\
			\\
			\\
			\\
			\\
			\\

		\end{tabular}
		\label{tbl:feature_correlation_IC}
	}

\vspace{-5mm}
\end{table}

\subsection{Premodel for Image Classification}

\subsubsection{Feature Selection} \label{sec:ic_features}
In this work, we considered a total of 29 candidate features, shown in Table~\ref{tbl:all_features}. The features were chosen based on
previous image classification work~\cite{hassaballah2016image}, \eg edge based features (more edges lead to a more complex image), as well
as intuition based on our motivation (Section~\ref{sec:motivation}), \eg contrast (lower contrast makes it harder to see image content).
Table~\ref{tbl:feature_correlation_IC} summarizes the features removed using correlation-based feature selection, leaving 17~features.
Next, we iteratively evaluated feature importance and performed a greedy search that reduced our feature count down to 7~features (see
Table~\ref{tbl:chosen_features_IC}). This process is described in Section~\ref{sec:feature_sel}.

\subsubsection{Feature Analysis} \label{sec:feature_analysis}
We now analyze the importance of each feature that was chosen during our feature selection process. We calculate feature importance by
first training a \premodel using all of our chosen features ($n$), and note the accuracy of our \premodel. In turn, we then remove each
feature, retraining and evaluating our \premodel on the remaining $n-1$ features, noting the drop in accuracy. We then normalize the values
to produce a percentage of importance for each feature. Figure~\ref{fig:feature_analysis}a shows the top 5 dominant features based on their
impact on our \premodel accuracy. It is clear our features hold a very similar level of importance, ranging between 18\% and 11\% for our
most and least important feature, respectively. The similarity of feature importance is an indication that each of our features is able to
represent distinct information about each image. All of which is important for the prediction task at hand.

\subsubsection{Creating The Premodel}
Applying our automatic approach to \premodel creation, described in Section~\ref{sec:model_desc}, resulted in implementing a multiple
classifier architecture consisting of a series of simple \NN models. We found that \NN has a quick prediction time (<1ms) and achieves a
high accuracy for this problem. Furthermore, we applied our Model Selection Algorithm (Section~\ref{sec:classifier_selection}) to determine
which \CNNs to be included in the \premodel. As we have explained in Section~\ref{sec:classifier_selection}, this process resulted in a
choice of: \mn{MobileNet\_v1\_100} for \textit{Model-1}, \mn{Inception\_v4} for \textit{Model-2}, and, finally, \mn{ResNet\_v1\_152} for
\textit{Model-3}. Finally, we use the training data generated in Section~\ref{sec:ic_features} and 10-fold-cross-validation to train and
evaluate our \premodel.

\begin{figure*}[t!]
	\centering
	\begin{tabularx}{1\textwidth} {>{\centering\arraybackslash}m{0.21\textwidth}>{\centering\arraybackslash}m{0.21\textwidth}>{\centering\arraybackslash}m{0.24\textwidth}>{\centering\arraybackslash}m{0.25\textwidth}}

		\includegraphics[width=0.210\textwidth,clip]{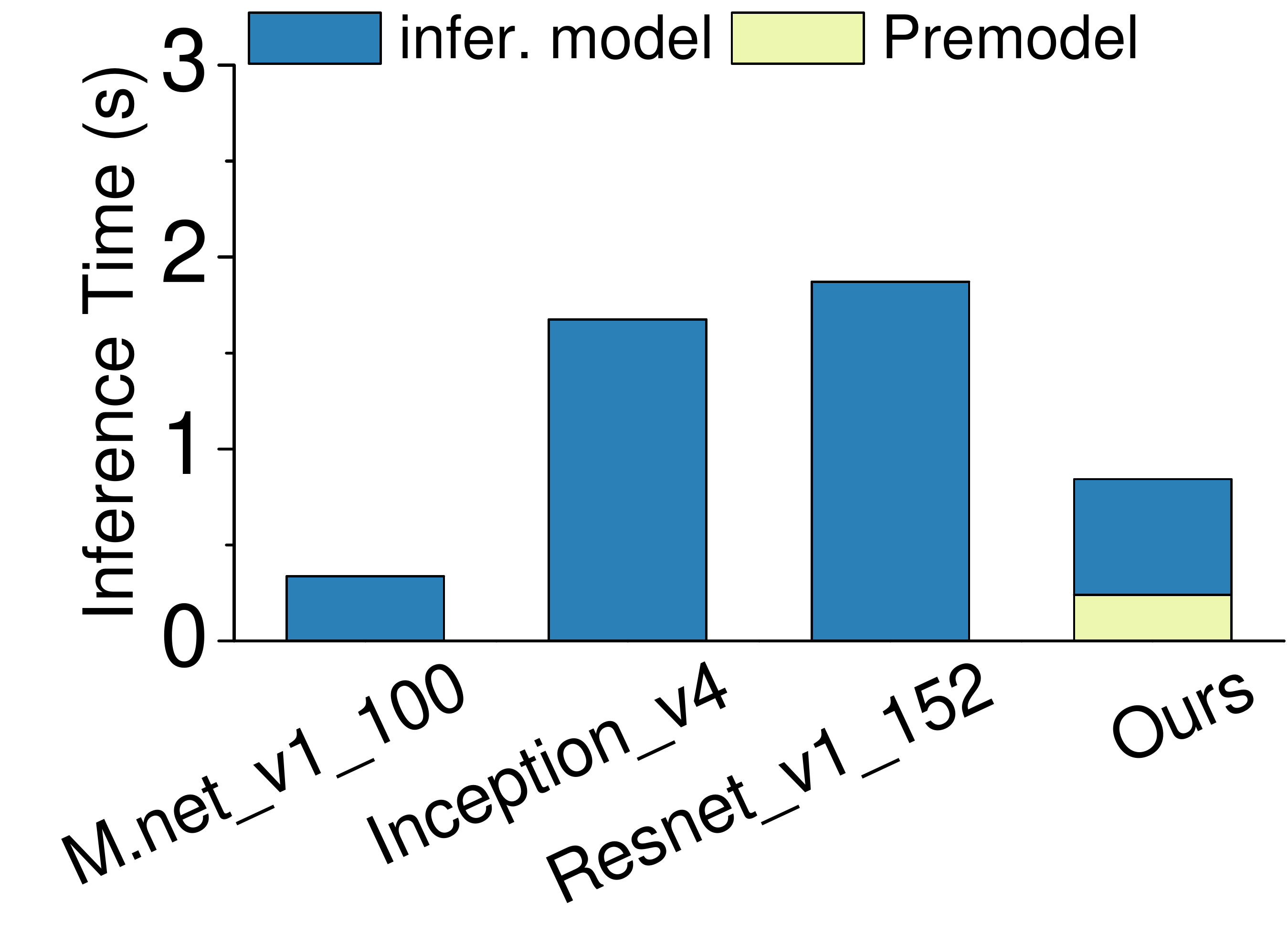} &
		\includegraphics[width=0.210\textwidth,clip]{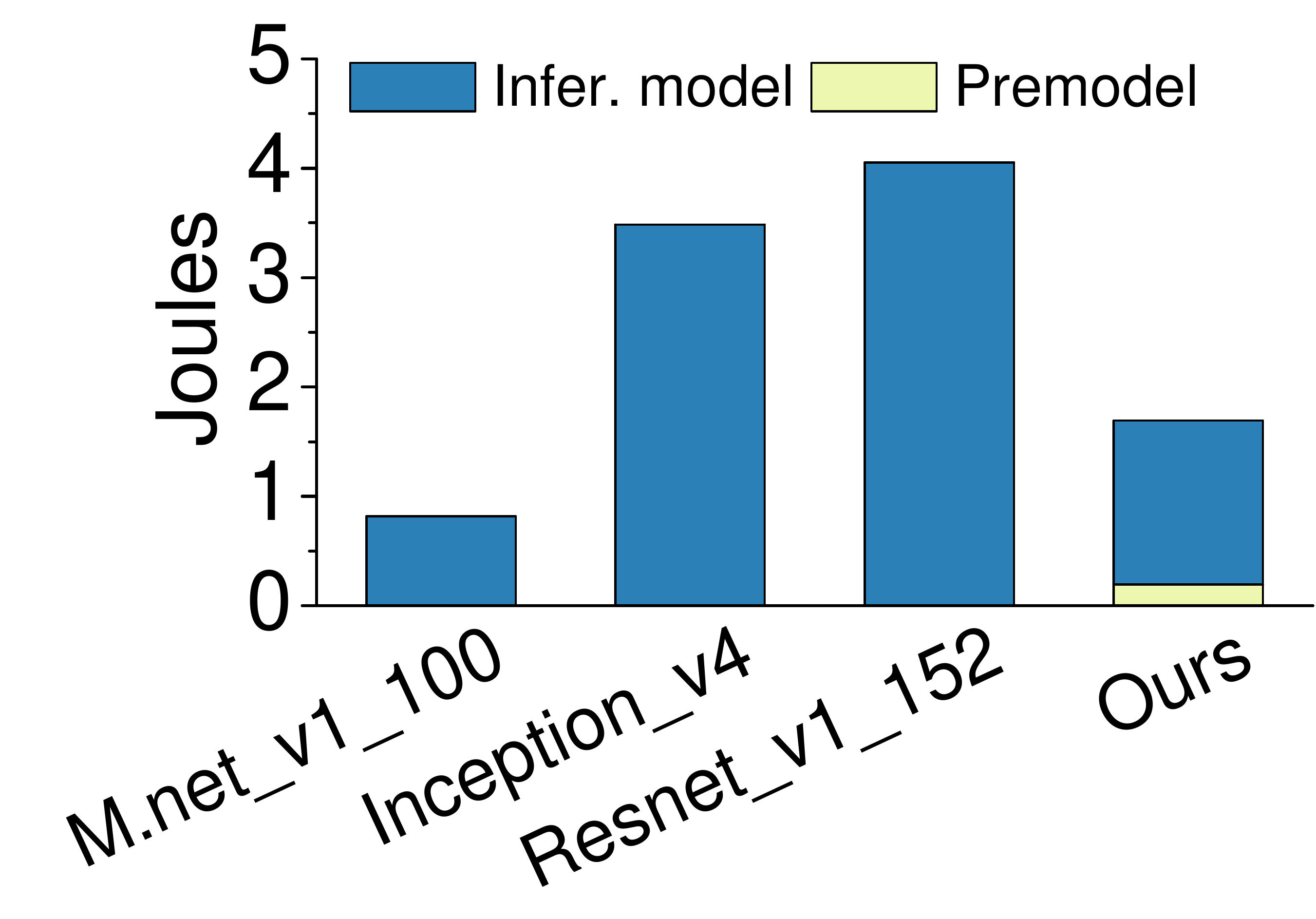} &	
		\includegraphics[width=0.240\textwidth,clip]{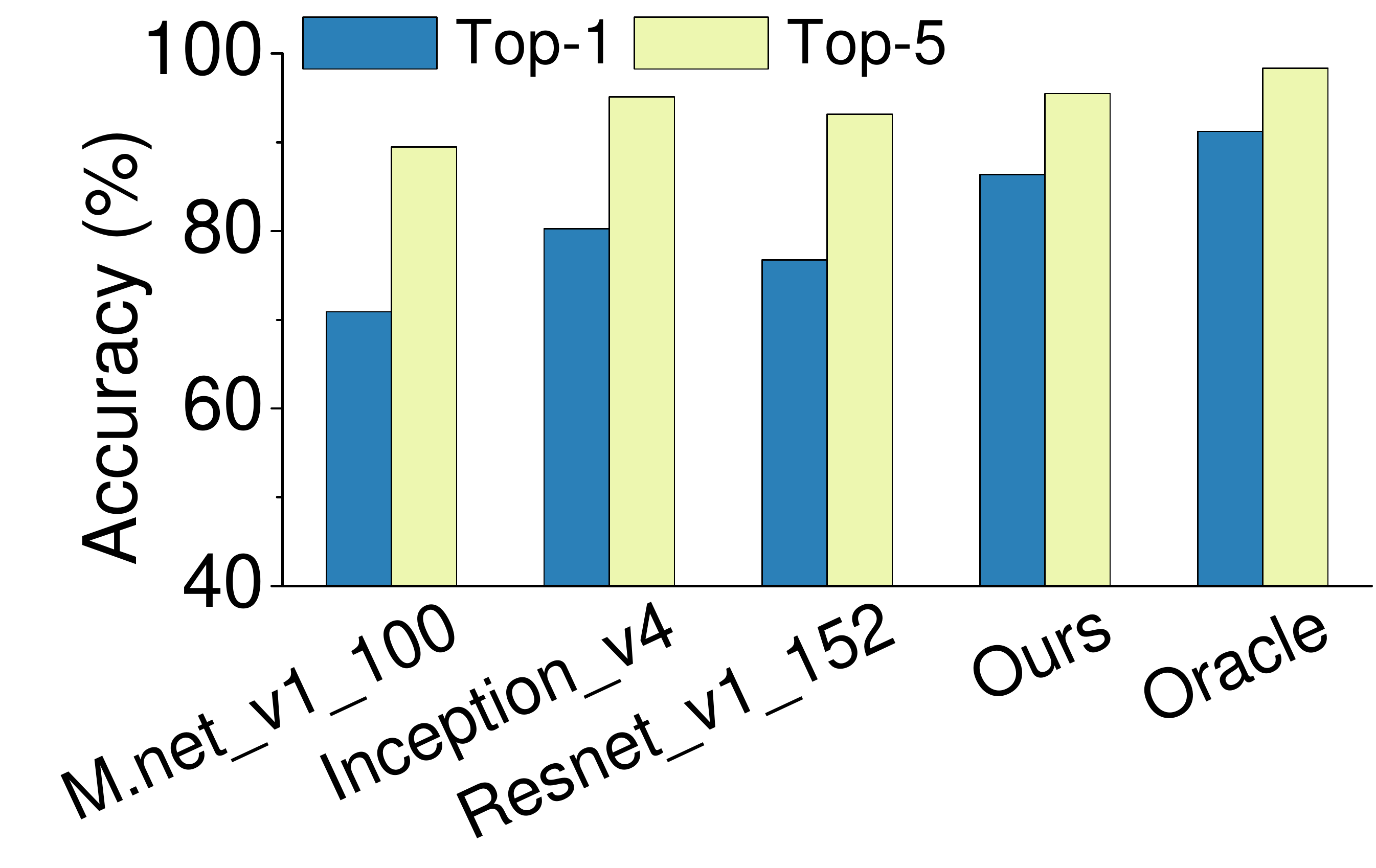} &
		\includegraphics[width=0.250\textwidth,clip]{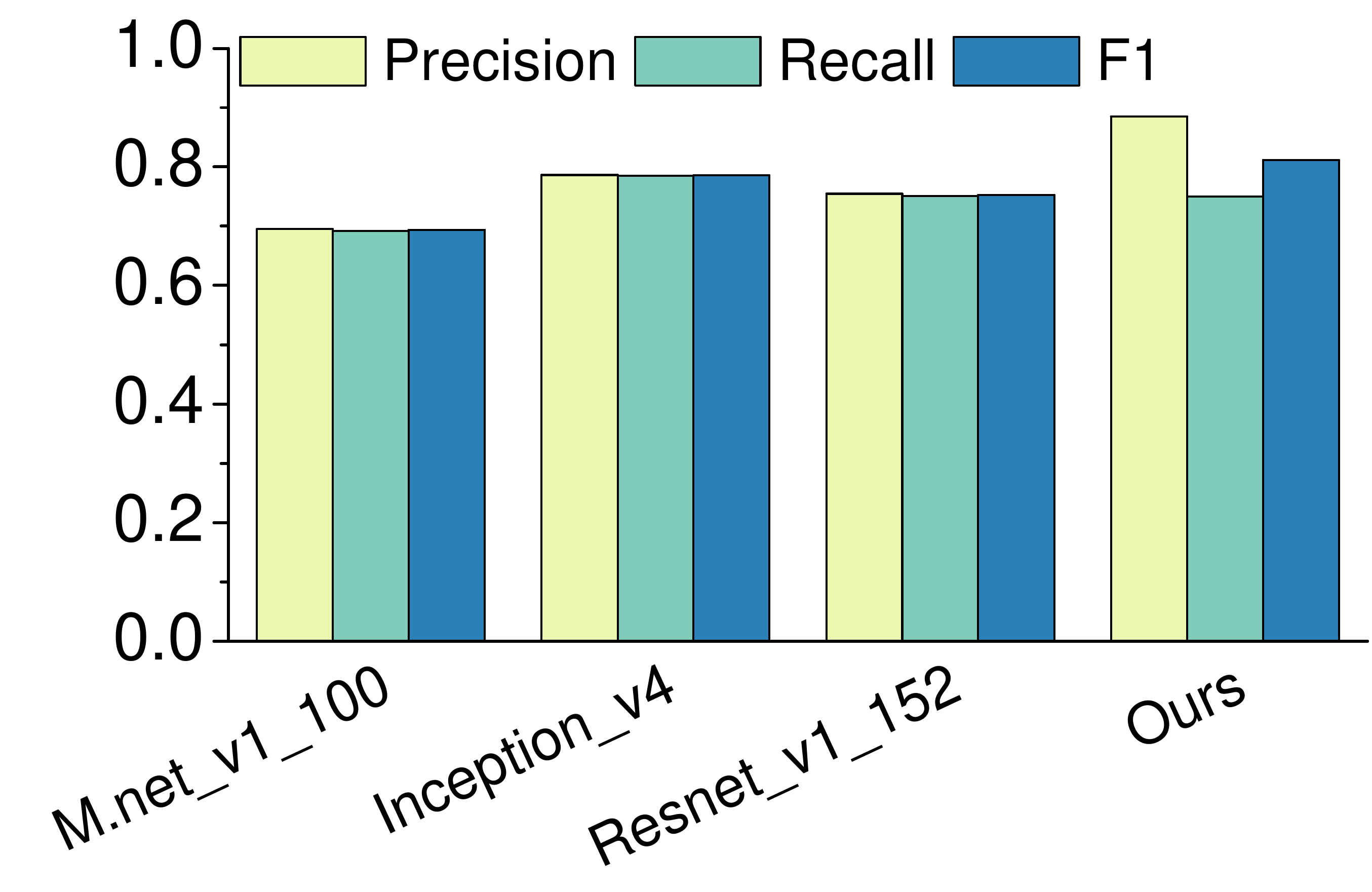} \\
		{\centering \scriptsize (a) Inference Time} &
		{\centering \scriptsize (b) Energy Consumption } &
		{\centering \scriptsize (c) Accuracy} 	&
		{\centering \scriptsize (d) Precision, Recall \& F1} \\
		
	\end{tabularx}
	\caption{
    Image Classification -- Overall performance of our approach against individual models and an \texttt{Oracle} for inference time (a), energy consumption (b), accuracy (c), and precision, recall and F1 scores (d).}
	\label{fig:expermiental_results}
\end{figure*}

\subsection{Overall Performance of Image Classification} 
\subsubsection{Inference Time} Figure~\ref{fig:expermiental_results}a compares the inference time among \DNN models used by our \premodel and our
approach.
Due to space limitations we limit to these three models (\texttt{MobileNet\_v1\_100}, \texttt{Inception\_v4}, and \texttt{ResNet\_v1\_152}) since they are the ones used by our \premodel.
\texttt{MobileNet\_v1\_100} is the fastest model for inferencing, being 2.8x and 2x faster than \texttt{Inception\_v4} and \texttt{ResNet\_v1\_152},
respectively, but is least accurate (see Figure~\ref{fig:expermiental_results}c).
The average inference time of our approach is under a second, which is
slightly longer than the 0.4 second average time of \texttt{MobileNet\_v1\_100}.
Our slower time is a result of using a \premodel, and choosing \texttt{Inception\_v4} or \texttt{ResNet\_v1\_152} on occasion.
Most of the overhead of our \premodel comes from feature extraction.
Our approach is 1.8x faster than \texttt{Inception\_v4}, the most
accurate inference model in our model set. Given that our approach can significantly improve the prediction accuracy of \texttt{MobileNet\_v1\_100},
we believe the modest cost of our \premodel is acceptable.

\subsubsection{Energy Consumption}
Figure~\ref{fig:expermiental_results}b gives the energy consumption. On the Jetson TX2 platform, the
energy consumption is proportional to the model inference time. As we speed up the overall inference, we reduce the energy consumption by
more than 2x compared to \texttt{Inception\_v4} and \texttt{ResNet\_v1\_152}. The energy footprint of our \premodel is small, being 4x and 24x lower
than \texttt{MobileNet\_v1\_100} and \texttt{ResNet\_v1\_152} respectively. As such, it is suitable for power-constrained devices, and can be used to
improve the overall accuracy when using multiple inferencing models. Furthermore, in cases where the \premodel predicts that none of the
\DNN models can successfully infers an input, it can skip inference to avoid wasting power. It is to note that since our \premodel runs on
the CPU, its energy footprint ratio is smaller than that for runtime.

\subsubsection{Accuracy}
Figure~\ref{fig:expermiental_results}c compares the \topone and \topfive accuracy achieved by each approach. We
also show the best possible accuracy given by a \emph{theoretically} perfect predictor for model selection, for which we call
\texttt{Oracle}. Note that the \texttt{Oracle} does not give a 100\% accuracy because there are cases where all the \DNNs fail.
However, not all \DNNs fail on the same images, \ie \texttt{ResNet\_v1\_152} will successfully classify some images which
\texttt{Inception\_v4} will fail on.
Therefore, by effectively leveraging multiple models, our approach outperforms all individual inference models.
It improves the accuracy of
\texttt{MobileNet\_v1\_100} by 16.6\% and 6\% for the \topone and the \topfive scores, respectively. It also improves the \topone accuracy of
\texttt{ResNet\_v1\_152} and \texttt{Inception\_v4} by 10.7\% and 7.6\%, respectively. While we observe little improvement for the \topfive score over
\texttt{Inception\_v4} -- just 0.34\% -- our approach is 2x faster than it. Our approach delivers over 96\% of the \texttt{Oracle} performance
(86.3\% vs 91.2\% for \topone and 95.4\% vs 98.3\% for \topfive).
This shows that our approach can improve the inference accuracy of individual models.
Overall, we achieve a 7.52\% improvement in accuracy over the most-capable single \DNN model, while reducing inference time by 1.8x.

\subsubsection{Precision, Recall, F1 Score}
Finally, Figure~\ref{fig:expermiental_results}d shows our approach outperforms individual \DNN
models in other evaluation metrics. Specifically, our approach gives the highest overall precision, which in turns leads to the best F1
score. High precision can reduce false positive, which is important for certain domains like video surveillance because it can reduce the
human involvement for inspecting false positive predictions.


\begin{table}[t!]
    \parbox{.45\linewidth}{
        \small
        \centering
        \captionsetup{justification=centering}
        \caption{Image Classification -- The final chosen features after feature selection.}
        \vspace{-3mm}
        \midsepremove
        \begin{tabular}{lll}
            \toprule
            \rowcolor{Gray}     \textit{n\_keypoints}           & \textit{avg\_perc\_brightness}   & \textit{hue1}              \\
                                \textit{contrast}               & \textit{area\_by\_perim}              & \textit{edge\_length1}     \\
            \rowcolor{Gray}     \textit{aspect\_ratio}          &                                       &                            \\
            \bottomrule
        \end{tabular}
        \label{tbl:chosen_features_IC}
    }
    \hfill
    \parbox{.45\linewidth}{
        \small
        \centering
        \captionsetup{justification=centering}
        \caption{Machine Translation -- The final chosen features after feature selection.}
        \midsepremove
        \begin{tabular}{l}
            \toprule
            \rowcolor{Gray}     \textit{n\_words}     \\
                                \textit{avg\_adj}     \\
            \rowcolor{Gray}     \textit{BoW}     \\
            \bottomrule
        \end{tabular}
        \label{tbl:chosen_features_MT}
    }
\end{table}

\begin{figure*}[t!]
	\centering
	\begin{tabularx}{1\textwidth} {>{\centering\arraybackslash}m{0.45\textwidth}>{\centering\arraybackslash}m{0.45\textwidth}}

		\includegraphics[width=0.42\textwidth,clip]{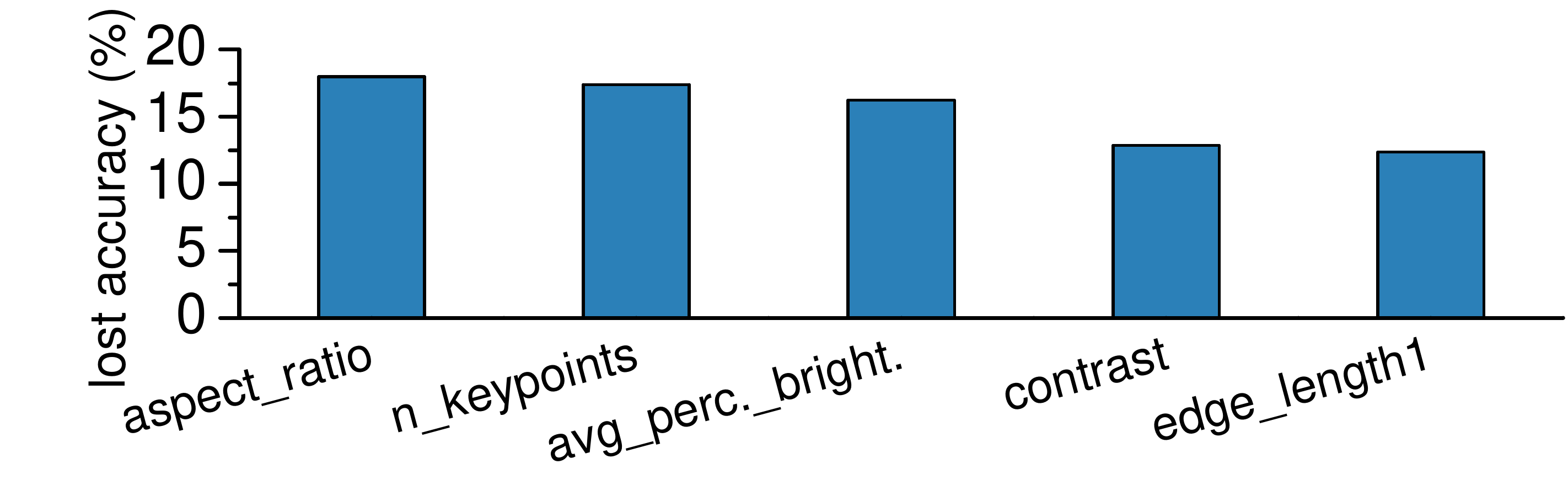} &
		\includegraphics[width=0.42\textwidth,clip]{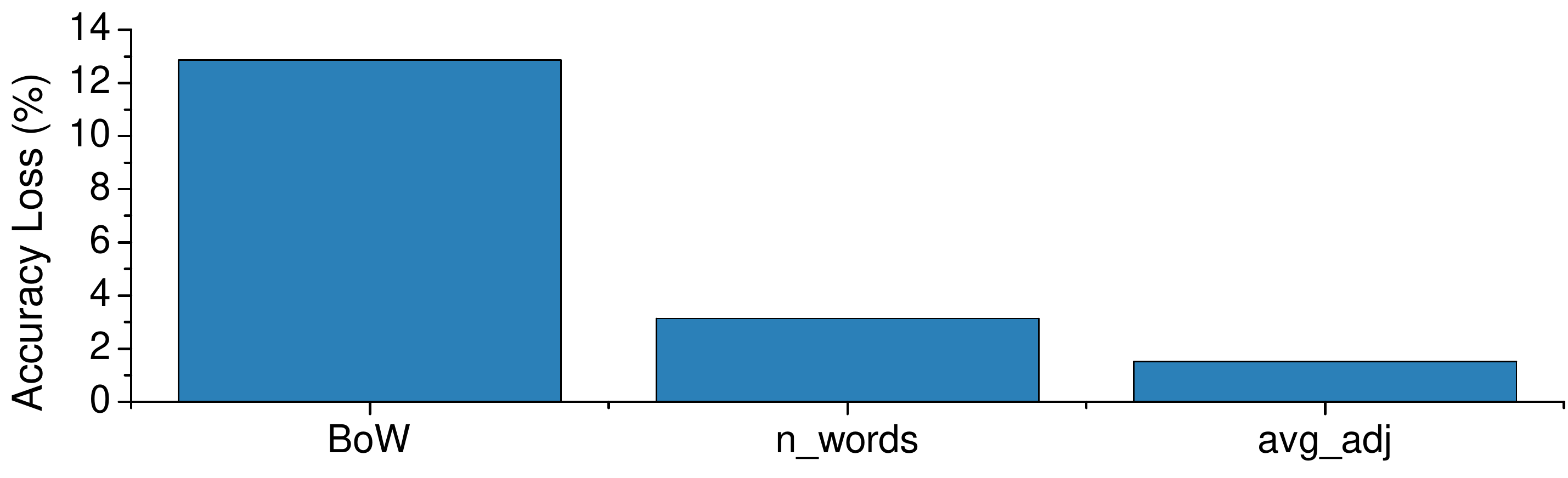} \\
		{\centering \scriptsize (a) Image Classification} &
		{\centering \scriptsize (b) Machine Translation} 	\\
	\end{tabularx}
	\caption{The loss in accuracy when final chosen features are not used in our \premodel. For image classification (a) we only show the top five. For machine translation (b) we show all 3.}
	\label{fig:feature_analysis}
\end{figure*}


\section{Case Study 2: Machine Translation} \label{sec:case2}
To evaluate our approach for machine translation we consider 15 \DNN models. We include
models of varying sizes and architectures, all trained using Tensorflow-NMT, a Neural Machine Translation library
provided by Tensorflow~\cite{luong17}. We name our models using the following convention: \texttt{\{gnmt\_\}N\_layer},
we prefix the name with \texttt{gnmt\_}where the model uses the Google Neural Machine Translation
Attention~\cite{45610}, and $N$ is the number of layers in the model. \eg \texttt{4\_layer} is a default Tensorflow-NMT
model made up of 4 layers. The models were trained on the \emph{WMT09-WMT14} English-German newstest dataset, and we
use the \emph{WMT15/16} English-German newstest dataset~\cite{wmt} to create our \premodel training data. Using
\emph{10-fold-cross-validation} on our \premodel to give a end-to-end analysis of our approach.

\vspace{+2mm}
\begin{table}[t!]
	\parbox{.45\linewidth}{
		\small
		\centering
		\captionsetup{justification=centering}
		\caption{All features considered for machine translation. See Section~\ref{sec:features}}
		\vspace{-3mm}
		\def\arraystretch{0.92}
		\scriptsize
		\centering
		\midsepremove
		\begin{tabular}{ll}
			\toprule
			\textbf{Feature}                    & \textbf{Description}                                    \\
			\midrule
			\rowcolor{Gray}     \textit{n\_words}                   & \# words in the sentence                                \\
			\textit{n\_bpe\_chars}              & \# bpe characters in a sentence                         \\
			\rowcolor{Gray}     \textit{avg\_bpe}                   & Average number of bpe characters per word               \\
			\textit{n\_tokens}                  & \# tokens in the sentence when tokenized                \\
			\rowcolor{Gray}     \textit{avg\_noun}                  & Average number of nouns per word                        \\
			\textit{avg\_verb}                  & Average number of verbs per word                        \\
			\rowcolor{Gray}     \textit{avg\_adj}                   & Average number of adjectives per word                   \\
			\textit{avg\_sat\_adj}              & Average number of satellite adjectives per word         \\
			\rowcolor{Gray}     \textit{avg\_adverb}                & Average number of adverbs per word                      \\
			\textit{avg\_punc}                  & Average punctuation characters per word                   \\
			\rowcolor{Gray}     \textit{avg\_word\_length}          & Average number of characters per word                   \\
			\bottomrule
		\end{tabular}
		\label{tbl:all_features_nmt}
	}
	\hfill
	\parbox{.45\linewidth}{
		\small
		\centering
		\captionsetup{justification=centering}
		\caption{Correlation values (absolute) of removed features to the kept ones for machine translation.}
		\vspace{-3mm}
		\def\arraystretch{0.92}
		\scriptsize
		\centering
		\midsepremove
		\begin{tabular}{lll}
			\toprule
			\textbf{Kept Feature}                                   & \textbf{Removed Feature}  &\textbf{Correl.}       \\
			\midrule
			\rowcolor{Gray}                                         & \textit{n\_bpe\_chars}    & 0.96                  \\
			\rowcolor{Gray} \multirow{-2}{*}{\textit{n\_words}}     & \textit{n\_tokens}        & 0.99                  \\
			\midrule
			\\
			\\
			\\
			\\
			\\
			\\
			\\
			\\
			\\
			
		\end{tabular}
		\label{tbl:feature_correlation_MT}
	}
\vspace{-3mm}
\end{table}

\vspace{-2mm}
\subsection{Premodel for Machine Translation}

\subsubsection{Feature Selection} \label{sec:mt_features} We considered a total of 11 features, which can be seen in
Table~\ref{tbl:all_features_nmt}, and a Bag of Words (BoW) representation of each sentence (explained in more detail below). Similar to
image classification, we chose our candidate features based on previous work~\cite{lui2012feature, khoo2006experiments}, \eg BoW, as well
as intuition based on our motivation (Section~\ref{sec:motivation}), \eg \textit{n\_words} (longer sentences are more complex and require a
more complex translator).

\cparagraph{Bag of words.} Applying our method to machine translation brings with it the need to classify each sentence to predict the
optimal \DNN. Text classification is a notoriously difficult task, and is made more difficult when we only have a single sentence to gather
features from. We are able to create a successful \premodel only using the features described in Table~\ref{tbl:all_features_nmt}. However,
with the addition of a Bag of Words (BoW) representation of each sentence we saw an increase in accuracy. Furthermore, previous work in
sentence classification~\cite{lui2012feature,khoo2006experiments,Magdy2017catfishes} often use a BoW representation, suggesting that BoW
can be useful for characterizing and modeling a sentence. A BoW representation of text describes the occurrence of words within the text.
It is represented as a vector that is based on a vocabulary. We generated a domain specific vocabulary based on all words in our training
dataset. Finally, we used Chi-square (Chi2) to perform feature reduction, which is widely used for BoW, leaving us with a BoW feature
vector of length 1500. We include a full evaluation of the effect of BoW and Chi2 feature selection on our machine translation \premodel in
Section~\ref{sec:MT_feature_importance}.

Table~\ref{tbl:feature_correlation_MT} summarizes the features we removed during the first stage of feature selection, leaving 9~features.
During the second stage we reduced our
feature count down to 3~features (see Table~\ref{tbl:chosen_features_MT}).
Figure~\ref{fig:feature_analysis}b summarizes the accuracy loss by removing any of the three selected features;
the two shown in Table~\ref{tbl:chosen_features_MT}, and a BoW representation.
It can be seen that by including BoW we reach a much higher accuracy.
This is to be expected, as BoW is a well researched and used representation of text input.
If we remove either \textit{n\_words} or \textit{avg\_adj} there is a small drop in accuracy, this indicates that BoW is able to capture similar information.
We chose to keep both of these features as they bring a small increase to accuracy with negligible overhead.

\subsubsection{Creating The Premodel} Using our approach resulted in implementing a single \NB classifier \premodel. We believe that a
single architecture \premodel was chosen because of our reduced dataset, \ie we have one tenth of the training data compared to image
classification. \NB achieved a high accuracy for this task, and has a quick prediction time (<1ms).

Applying our Model Selection Algorithm, we set \textit{selection\_method} to `\textit{Accuracy}' and $\theta$ to 2.0. Again, see
Section~\ref{sec:msa_eval} for a sensitivity analysis of these parameters. This resulted in a \premodel selection of \mn{gnmt\_2\_layer},
\mn{gnmt\_8\_layer}, and \mn{gnmt\_3\_layer} for \textit{Model-1}, \textit{Model-2}, and \textit{Model-3}, respectively. Finally, we use
the training data generated in Section~\ref{sec:mt_features} and 10-fold-cross-validation to train and evaluate our \premodel.

\begin{figure*}[t!]
	\centering
	\begin{tabularx}{1\textwidth} {>{\centering\arraybackslash}m{1.6in}>{\centering\arraybackslash}m{1.6in}>{\centering\arraybackslash}m{1.6in}}
		
		\includegraphics[width=0.300\textwidth,clip]{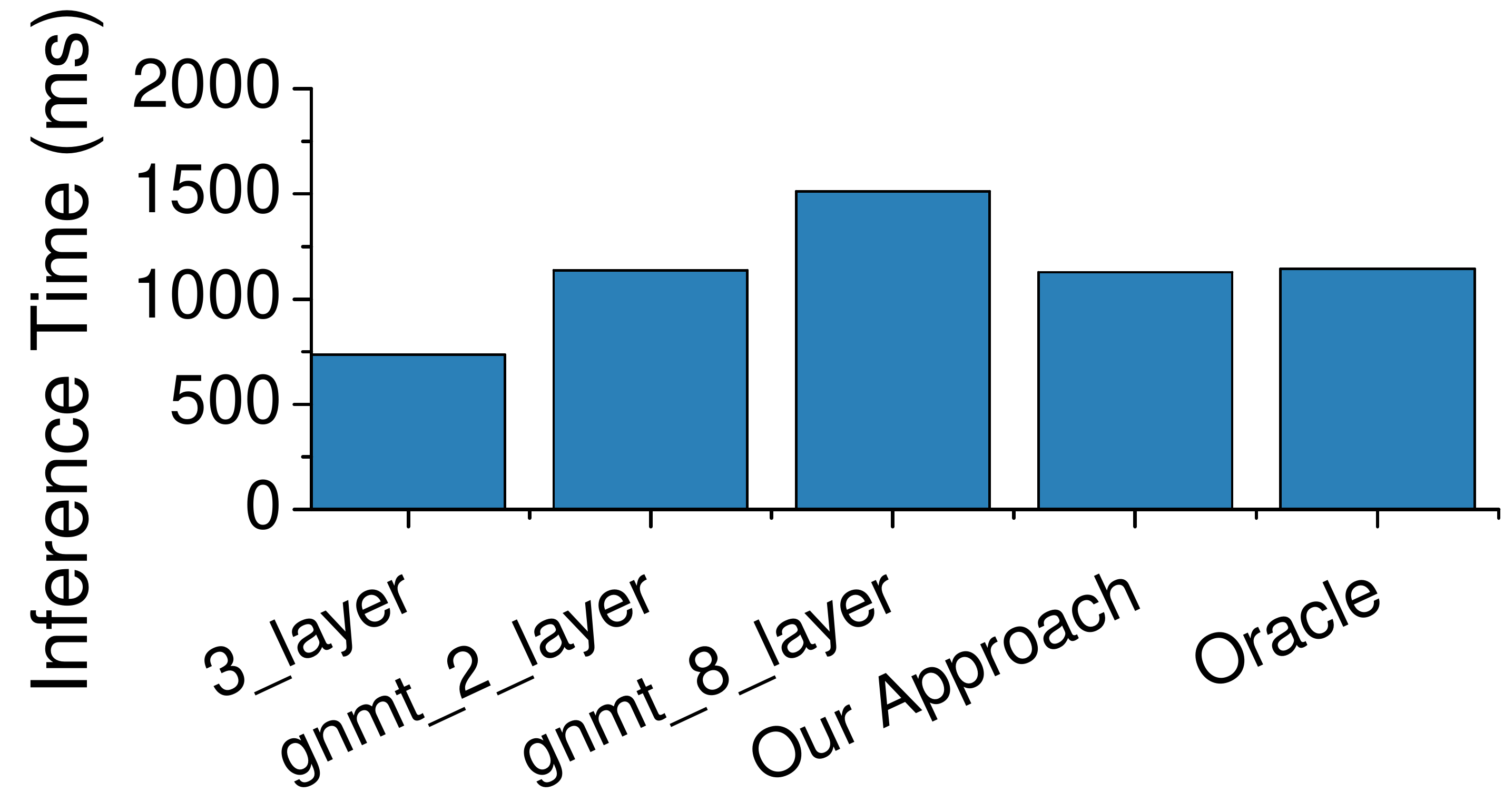} &
		\includegraphics[width=0.30\textwidth,clip]{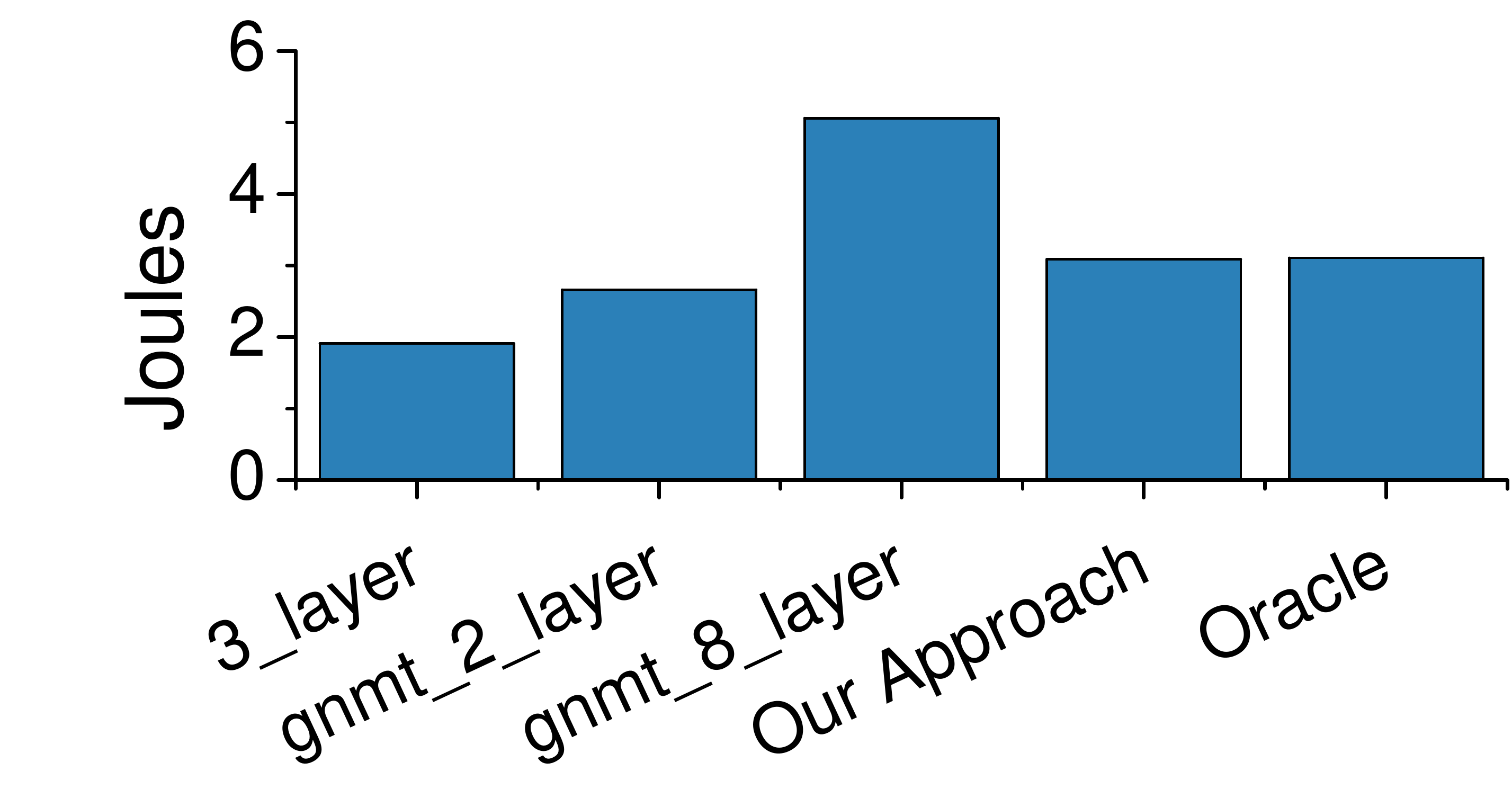} &	
		\includegraphics[width=0.30\textwidth,clip]{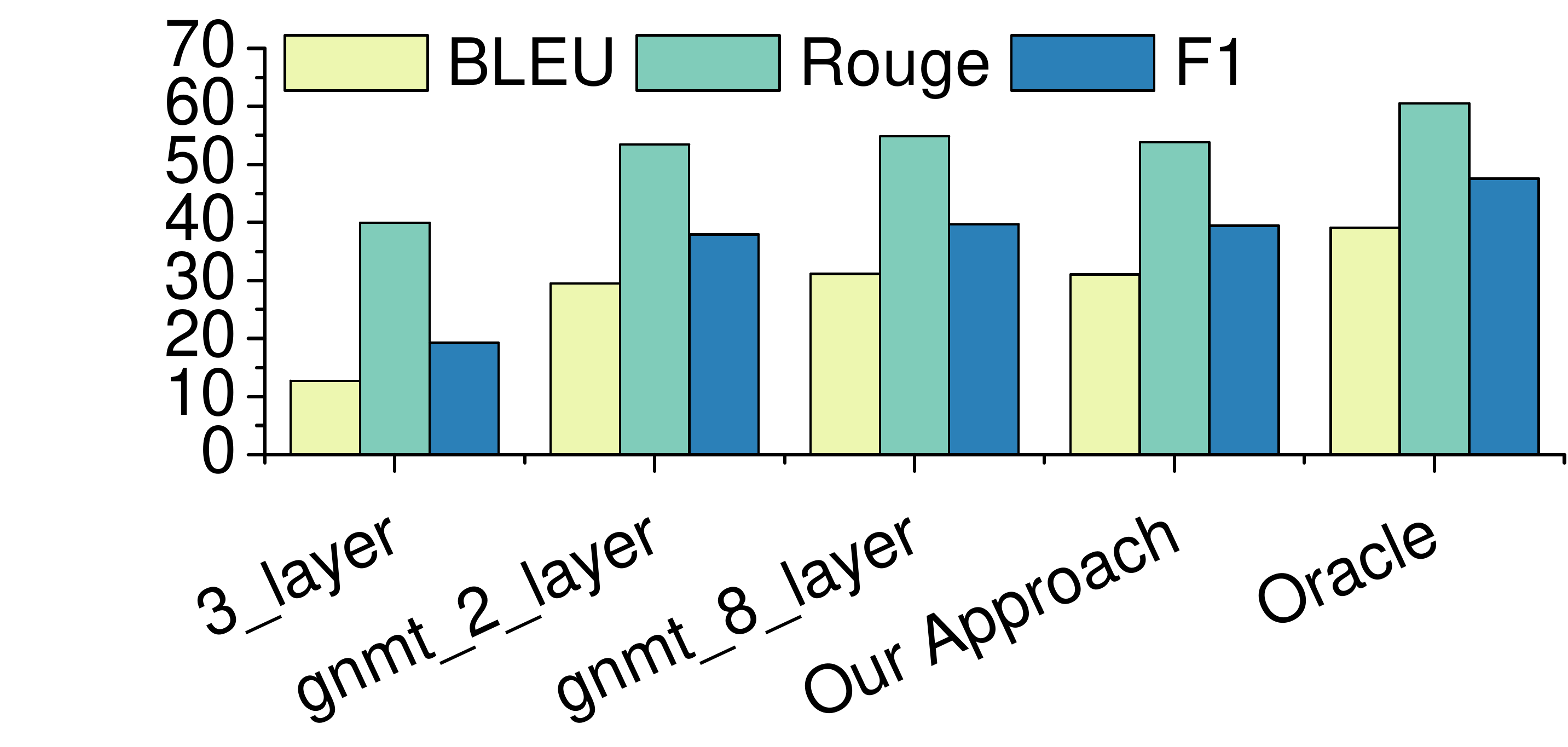} \\
		{\centering \scriptsize (a) Inference Time} &
		{\centering \scriptsize (b) Energy Consumption } &
		{\centering \scriptsize (c) \bleu, \rouge, and F1} 	\\
	\end{tabularx}
	\caption{Machine Translation -- Overall performance of our approach against individual models and an \texttt{Oracle}.}
	\label{fig:MT_all_results}
\end{figure*}

\subsection{Overall Performance for Machine Translation}

In this section, we evaluate our methodology when applied to Neural Machine Translation (\NMT). We compare our approach to three other \NMT
models considered in our \premodel. We chose these models as they show a range of complexity and capability. Furthermore, we compare our
approach to an \texttt{Oracle}, a \emph{theoretical} perfect approach that achieves the best possible score for each metric.

\subsubsection{Inference Time} As depicted in Figure~\ref{fig:MT_all_results}a, \texttt{3\_layer} is the quickest \DNN, 1.55x faster than
the \texttt{Oracle} and 2.05x faster than the most complex individual \DNN, \texttt{gnmt\_8\_layer}. However, \texttt{3\_layer} is also the
least accurate \DNN (Figure~\ref{fig:MT_all_results}c) as it is outperformed in every accuracy metric by all other approaches. Our
approach, the \oracle, and \texttt{gnmt\_2\_layer} have very similar inference times; nonetheless, our approach and the \oracle outperform
\texttt{gnmt\_2\_layer} for accuracy. The runtime of our \premodel and feature extraction is small, consisting of <1ms for the \premodel
and <5ms for feature extraction, per sentence. Feature extraction and \premodel overheads are included in the inference time of our
approach and the \oracle. Incidentally, our approach is slightly quicker than the \oracle; this is a result of our \premodel often
mispredicting \texttt{gnmt\_2\_layer} for \texttt{gnmt\_8\_layer} and vice versa. This specific misprediction makes up 38.5\% of the cases
where \premodel makes an incorrect prediction. To improve accuracy we will need more data to train our \premodel, as we currently have a
high feature to sentence ratio. Alternatively, we could deeply investigate the sentences that are best for each model and intuitively add a
new feature to our \premodel, however, the differences may not be intuitive to spot. Overall, we are 1.34x faster than the single most
capable \DNN without a decrease in accuracy.

\subsubsection{Energy Consumption.} Figure~\ref{fig:MT_all_results}b compares energy consumption, including \premodel costs, which are negligible (See Section~\ref{sec:overhead}).
Much like the image classification \DNNs, energy consumption is proportional to model inference time; therefore, as we reduce overall inference time we also improve energy efficiency.
A major difference between energy consumption and inference time is the emphasized ratios between each model,
\eg \texttt{gnmt\_2\_layer} is 1.24x quicker than \texttt{gnmt\_8\_layer}, but it uses 1.90x less energy, nearly half as much.
Overall, we use 1.39x less energy on average than the single most capable model, without a significant change in F1 measure.
Therefore, our methodology can be used to improve energy efficiency while having little impact on accuracy, or in some cases, seeing an improvement in accuracy.
Furthermore, our \premodel is able to predict when none of the \DNNs are able to give a suitable output, in this case we can skip inference to avoid wasting power.
Implementing this results in using 1.48x less energy on average than \texttt{gnmt\_8\_layer}.

\subsubsection{\bleu, \rouge, and F1 Measure} Figure~\ref{fig:MT_all_results}c compares \DNNs across our accuracy metrics.
We will mostly compare F1 measure here, but all metrics follow the same pattern.
As all models do not fail on the same sentences, we are able to achieve an overall better F1 measure by leveraging multiple \DNNs.
This can be seen by looking at the \oracle, which achieves an F1 measure of 47.54, a ~20\% increase over \texttt{gnmt\_8\_layer}, which achieves 39.71.
For this case study, we achieved 83\% of the \oracle F1 measure.
Overall our approach achieves approximately the same F1 measure as the single most capable model, and improves upon the accuracy of \texttt{gnmt\_2\_layer}
(the closest single \DNN in terms of inference time), by 4\%.
For our \premodel to achieve its full potential, as show by the \oracle, we require more data to train and test our \premodel.

\section{Analysis \label{sec:results}}
We now analyze the working mechanism of our approach to justify our design choices.

\vspace{-2mm}
\subsection{Alternative Techniques for Premodel} \label{sec:alt_premodels}
\begin{figure*}[t!]
	\centering
	\begin{tabularx}{1\textwidth} {>{\centering\arraybackslash}m{0.53\textwidth}>{\centering\arraybackslash}m{0.43\textwidth}}

		\includegraphics[width=0.560\textwidth,clip]{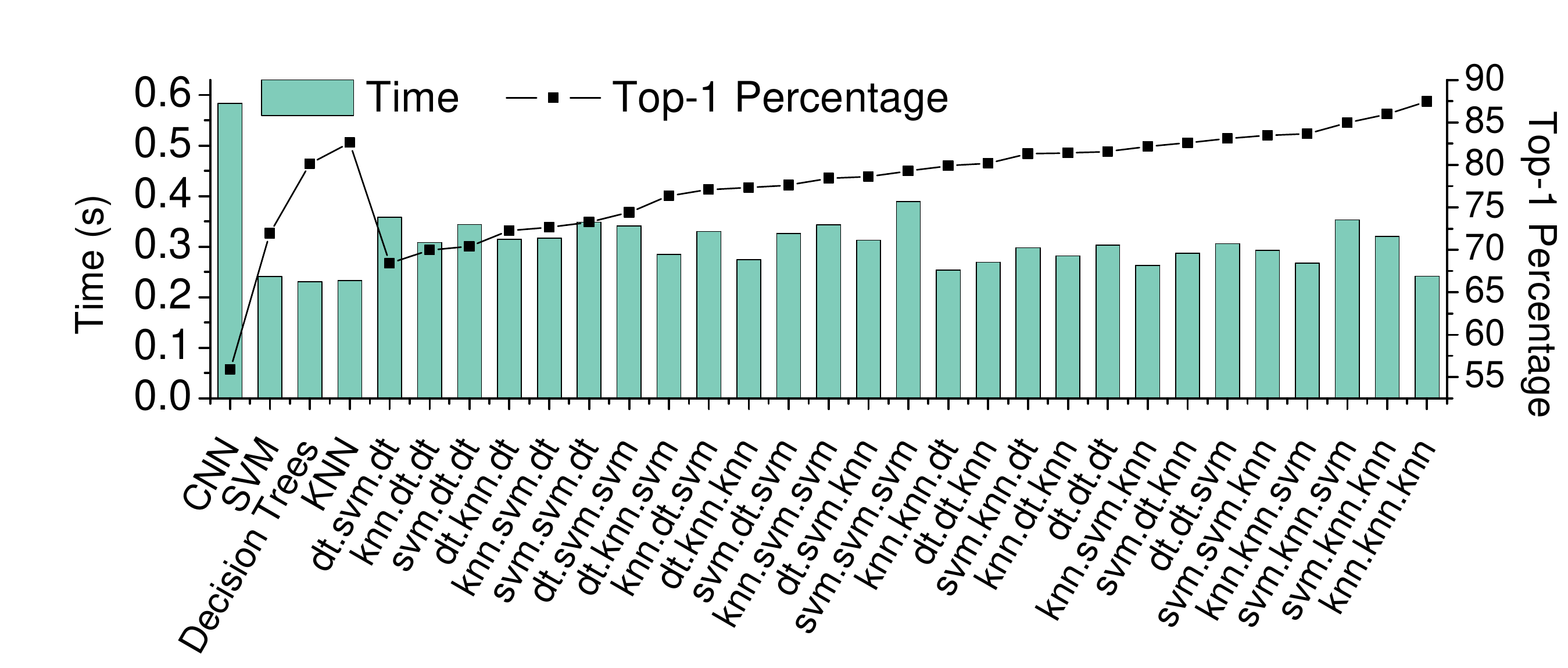} &
		\includegraphics[width=0.400\textwidth,clip]{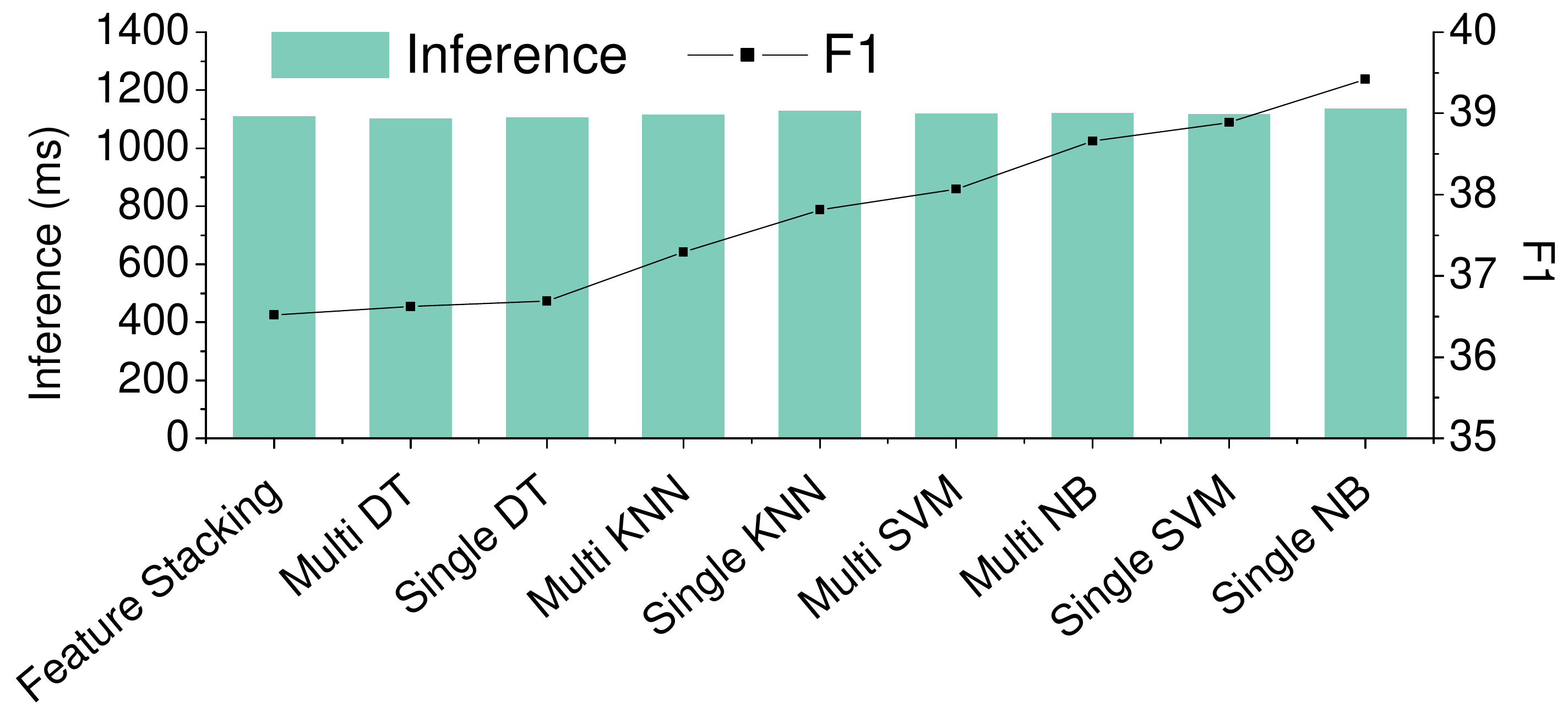} \\
		{\centering \scriptsize (a) Image Classification} &
		{\centering \scriptsize (b) Machine Translation} 	\\
	\end{tabularx}
	\caption{Comparison of alternative predictive modeling techniques for building the \premodel.}
	\label{fig:alternative_techniques}
\end{figure*}

\subsubsection{Image Classification} \label{sec:alt_premodel_IC}
Figure~\ref{fig:alternative_techniques}a shows the \topone accuracy and runtime for using different techniques to construct the \premodel.
Here, the learning task is to predict which of the inference models, \texttt{MobileNet}, \texttt{Inception}, and \texttt{ResNet}, to use.
In addition to our multi-classifier architecture made up of only \NN classifiers, we have considered different variations of Decision Trees
(\DT) and Support Vector Machines (\SVM). We also consider a single architecture \premodel using the above mentioned \ML techniques, and a
\CNN. Our \CNN-based \premodel is based on the MobileNet structure, which is designed for embedded inference. We train all models using the
same examples. We also use the same feature set for \NN, \DT, and \SVM. For the \CNN, we use an automated hyper-parameter tuner
\cite{klein2016fast} to optimize the training parameters, and we train the model for over 500 epochs.

\cparagraph{Notation.} In this instance, our multiple classifier architecture requires 3 components. We denote a \premodel configuration as
$X.Y.Z$ (see also Section~\ref{sec:multi_architecture}), where $X$, $Y$ and $Z$ indicate classifier for the first, second and third level
of the \premodel, respectively. For example, \NN.\SVM.\NN denotes using a \NN model for the first and last levels, with a \SVM model at the
second level.

While we hypothesized a \CNN model to be effective, the results are disappointing given its
high runtime overhead.
A \NN model has an overhead that is comparable to the \DT and the \SVM, but has a higher accuracy.
It is clear that our chosen \premodel architecture (\NN.\NN.\NN) was the best choice, it achieves
the highest \topone accuracy (87.4\%) and the fastest running time (0.20~second).
One of the benefits of using a \NN model in all levels is
that the neighbouring measurement only needs to be performed once as the results can be shared among models in different levels; \ie
the runtime overhead is nearly constant if we use the \NN across all hierarchical levels.
The accuracy for each of our \NN models in our \premodel is 95.8\%, 80.1\%, 72.3\%, respectively.

\subsubsection{Machine Translation}
Figure~\ref{fig:alternative_techniques}b shows the F1 measure and inference time for different architectures of \premodel when applied to the machine translation problem.
In this instance, we are predicting whether to use  \mn{gnmt\_2\_layer}, \mn{gnmt\_8\_layer}, or \mn{gnmt\_3\_layer} for translating an input sentence.
Our \premodel can also predict that all these translators will fail, making a total of~4~labels to choose from.
We evaluated single and multiple architectures, across \NN, \DT, \SVM, and \NB classifiers.
For multiple classifier architectures we carried out a less exhaustive search compared to Section~\ref{sec:alt_premodel_IC};
we discovered that best performance was often achieved by using the same classifier for each component.
Finally, we compare an alternate approach named feature stacking~\cite{lui2012feature}.
Using feature stacking we split classification into two classifiers, one using the BoW features, the other using our remaining features, we then use a probability measure choose the predicted model.

For this problem we can see that the single classifier architecture always outperforms its multiple classifier alternative.
This is likely as a result of our high dimensional feature space, with a comparatively low training set.
Feature stacking also had a poor performance for this problem, in fact it performs worse than all other architectures, indicating that our features work better together.
Overall, there is little variance in the runtime of each approach, every model achieves a runtime between 1100ms and 1140 ms.
Our chosen approach, a single \NB classifier, achieves the highest F1 measure overall, with very similar runtime to all other approaches.

\begin{figure}
	\includegraphics[width=0.45\textwidth]{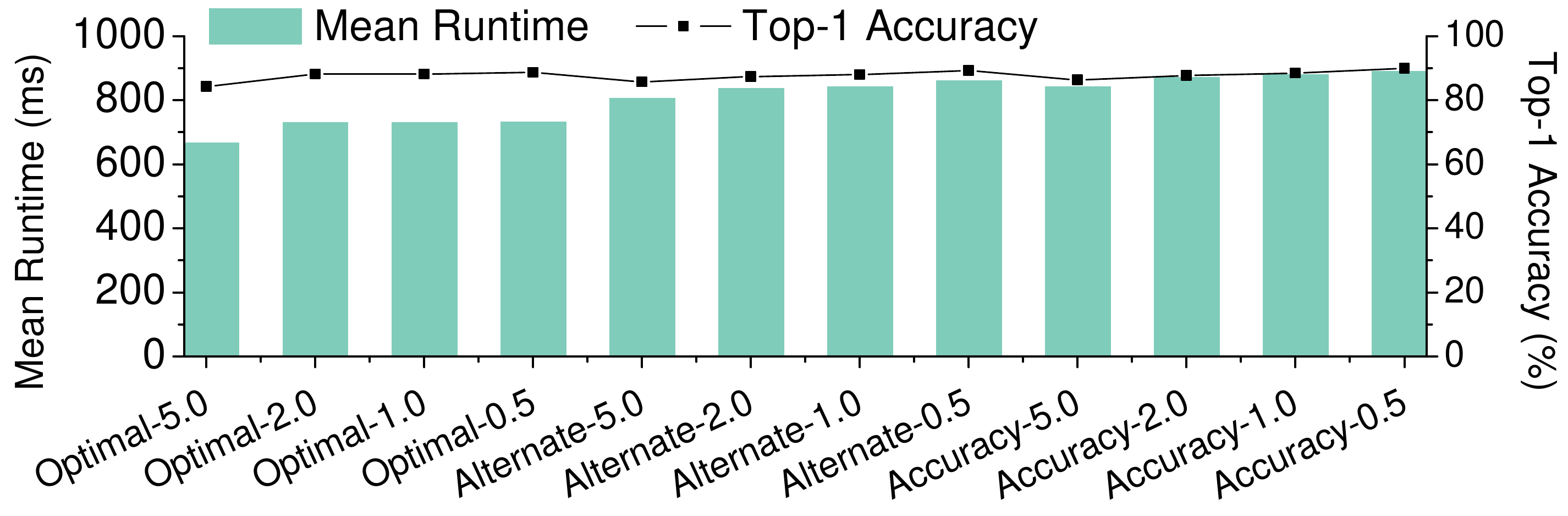}
	\caption{The inference time and Top-1 accuracy achieved when building a \premodel based on the Model Selection Algorithm configurations shown.}
	\label{fig:msa_configs}
\end{figure}

\subsection{Sensitivity Analysis for Model Selection Algorithm} \label{sec:msa_eval}

In Section~\ref{sec:classifier_selection}, we describe the algorithm we created to decide which \DNNs to include in our \premodel.
In this section we will analyze how changing the parameters given to the Model Selection Algorithm effect our \premodel, and the resultant end-to-end performance.
We will perform a case study using image classification, but the results for machine translation are very similar.
We consider the performance if we were able to create a perfect predictor as a \premodel,
this is to prevent our \premodel accuracy from introducing any noise and allowing us to evaluate the Model Selection Algorithm in isolation.
a total of 12 parameter configurations --  our three available choices for \textit{SelectionMethod} (defined in Section~\ref{sec:classifier_selection}), and 4 different choices for $\theta$ (5.0, 2.0, 1.0, and 0.5).
We take every combination of these parameters.

\cparagraph{Notation.} Our parameter configuration is \textit{SelectionMethod-$\theta$},
where \textit{SelectionMethod} is either \textit{Accuracy}, \textit{Optimal}, or \textit{Alternate};
and $\theta$ is our threshold parameter.
For example, the notation \textit{Accuracy-5.0}, means we always select the most accurate model
in each iteration of our algorithm, and we stop once our accuracy improvement is less than 5.0\%.

\subsubsection{Results}
Figure~\ref{fig:msa_configs} shows the effect of different parameters on our final \premodel results. As we decrease~$\theta$, our Model
Selection Algorithm will select more models to include in our \premodel, \eg The \premodel of \textit{Alternate-5.0} to
\textit{Alternate-0.5} is made up of 3, 4, 5, and 7 \DNN classifiers, respectively. Including more \DNNs results in a higher overall
\topone accuracy, however there are also some drawbacks. More \DNNs means more classes for our \premodel to choose between, therefore
making the job of the \premodel harder. It also means that we need to hold more \DNNs in memory, which could be an issue for devices with
limited memory (We discuss resource usage in more detail in Section~\ref{sec:resource_use}). It is worth noting that there is no change in
\DNN selection from \textit{Optimal-2.0} to \textit{Optimal-1.0}, as the next model that we can add only brings an accuracy improvement of
0.488.

Finally, we can see that each \textit{SelectionMethod} has its own 'profile', that is, each has its own positive and negative impact.
Figure~\ref{fig:msa_configs} shows that \textit{Optimal} results in an overall faster runtime, however, it has a lower \topone accuracy.
\textit{Accuracy} is able to achieve the highest possible \topone score, but this comes at the cost of speed, achieving ~1.26x slowdown for a 2\% accuracy increase.
\textit{Alternate} attempts to find a balance between the other two approaches, it is able to achieve and accuracy and runtime in between \textit{Optimal} and \textit{Accuracy}.

\subsection{Feature Importance} \label{sec:feature_importance}

\subsubsection{Image Classification}
\begin{figure}
	\centering
	\begin{minipage}{.47\textwidth}
		\centering
		\includegraphics[width=.95\linewidth]{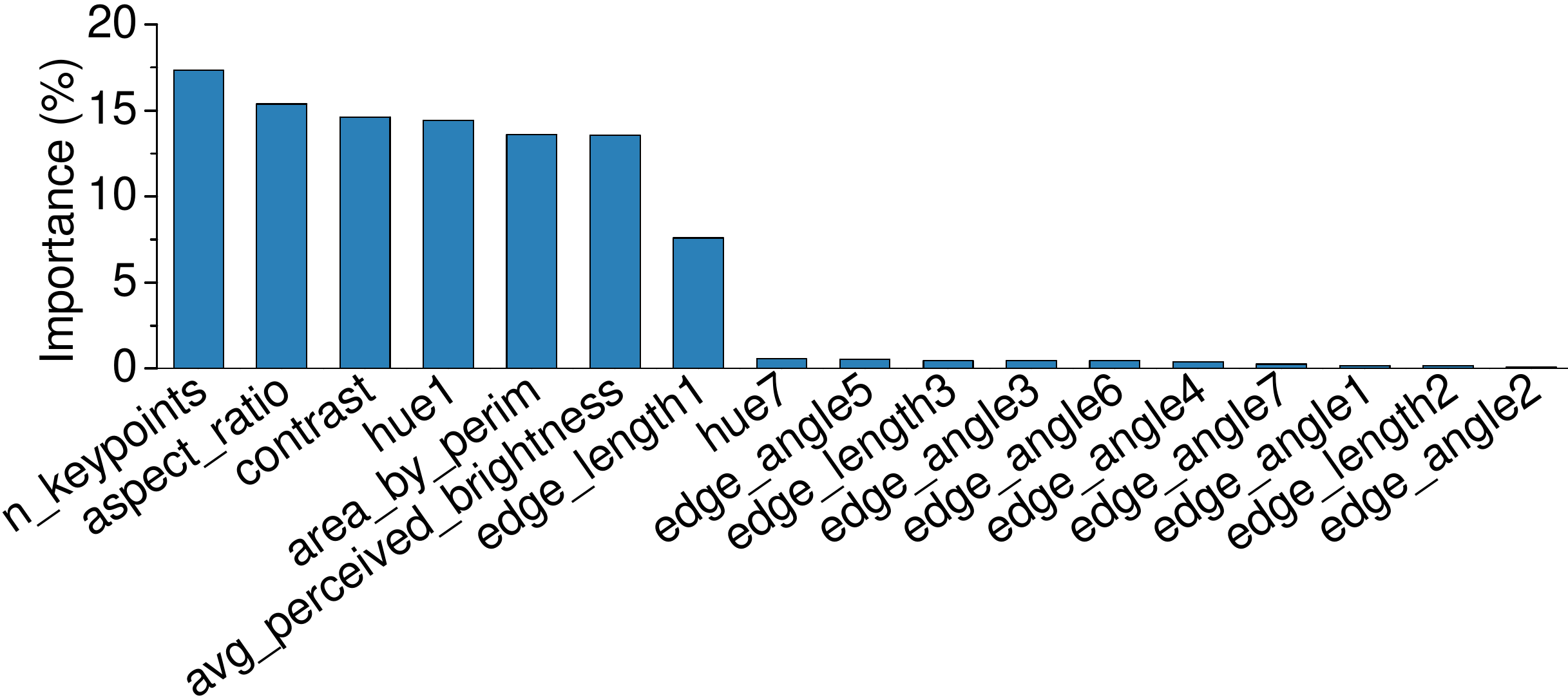}
		\caption{Image Classification -- Accuracy loss if a feature is not used.}
		\label{fig:feature_17_importance}
	\end{minipage}%
	\hfill
	\begin{minipage}{.47\textwidth}
		\centering
		\includegraphics[width=.95\linewidth]{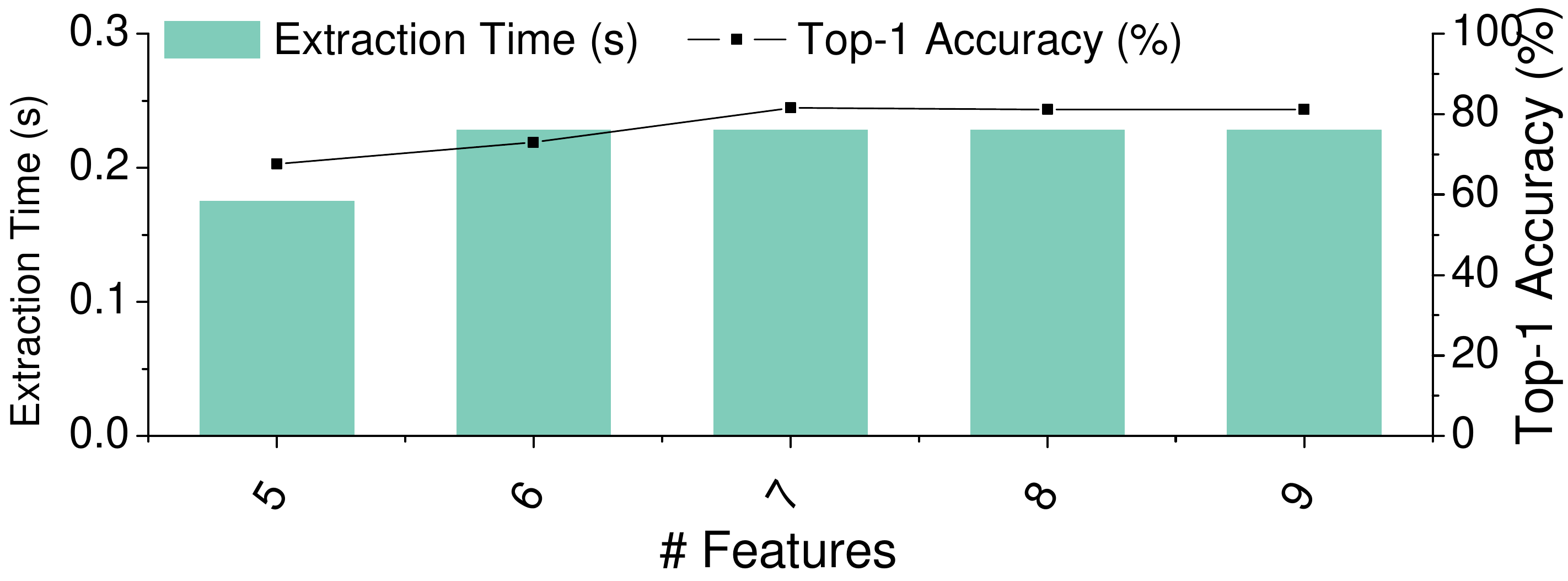}
		\caption{Image Classification -- The impact of \premodel feature count on \premodel runtime and overall \topone score.}
		\label{fig:feature_removal}
	\end{minipage}
\end{figure}

Our feature selection process (described in Section~\ref{sec:features}) resulted in using 7 features to represent each image to our
\premodel. In Figure~\ref{fig:feature_17_importance} we show the importance of all of the chosen features along with other considered ones
(given in Tables~\ref{tbl:all_features} and \ref{tbl:feature_correlation_IC}). The first 7 chosen features are the most important; there is
a sudden drop in feature importance at feature 8 (\textit{hue7}). Furthermore, Figure~\ref{fig:feature_removal} shows the impact on
\premodel execution time and \topone accuracy when we change the number of features used. By decreasing the number of features there is a
dramatic decrease \topone accuracy, with very little change in extraction time. To reduce overhead, we would need to reduce our feature
count to 5, however this comes at the cost of a 13.9\% decrease in \topone accuracy. By increasing the feature count it can be seen that
there is minor changes in overhead, but, surprisingly, there is actually also a small decrease in \topone accuracy of 0.4\%. From this we
can conclude that using 7 features is ideal.

\begin{figure}
\centering
\begin{minipage}{.47\textwidth}
  \centering
  \includegraphics[width=.95\linewidth]{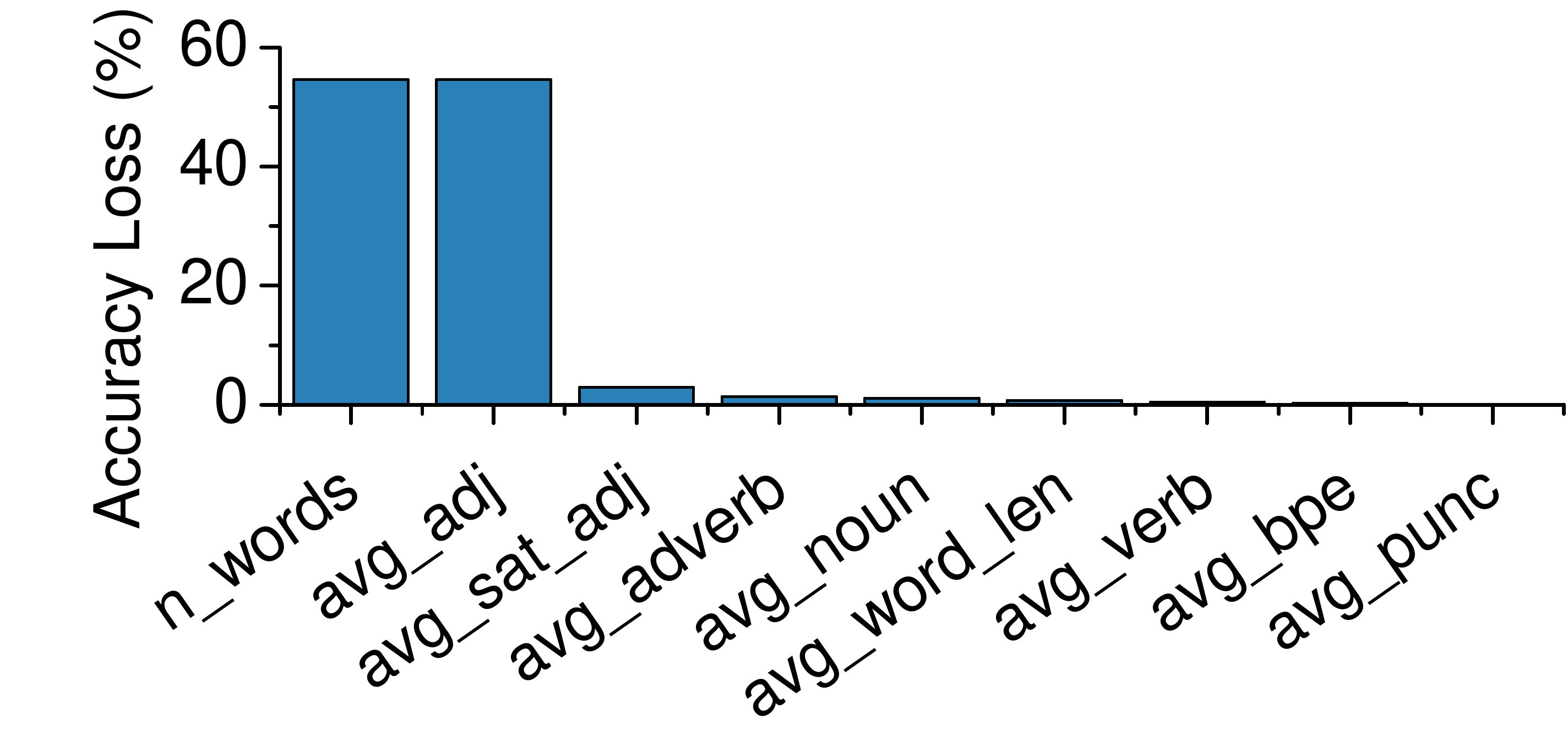}
  \caption{Machine Translation -- Accuracy loss if a~certain feature is not used.}
  \label{fig:MT_feature_selection}
\end{minipage}%
\hfill
\begin{minipage}{.47\textwidth}
  \centering
  \includegraphics[width=.95\linewidth]{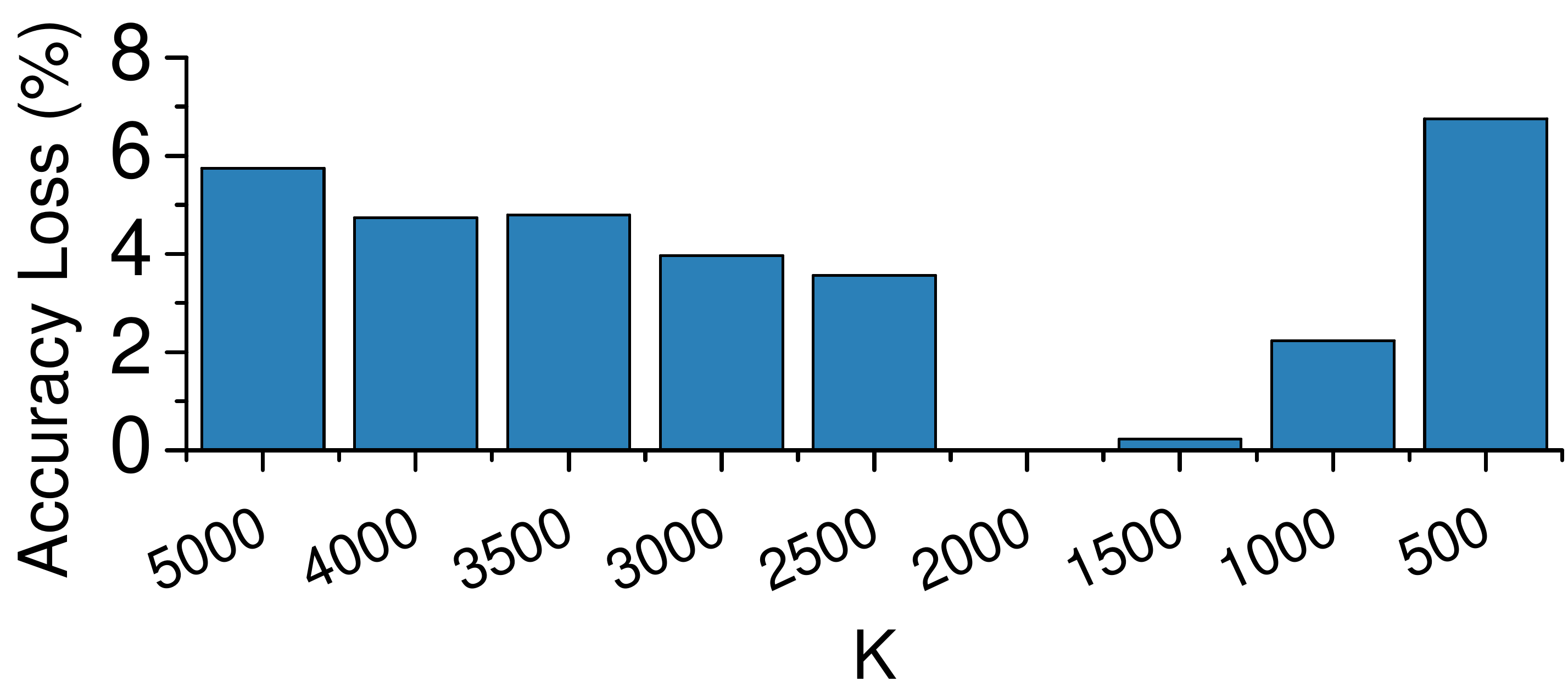}
  \caption{Machine Translation -- Accuracy loss for different values of $k$ using Bag of Words. $k$=2000 is our baseline.}
	\label{fig:bow_feature_selection}
\end{minipage}
\end{figure}

\subsubsection{Machine Translation} \label{sec:MT_feature_importance}
As with image classification above, in this section we will evaluate our feature selection process on the machine translation problem.
We will evaluate our BoW feature separately so clearly show the importance of all feature choices.
Figure~\ref{fig:MT_feature_selection} shows the importance of all our features which were not removed during our correlation check (See Table~\ref{tbl:feature_correlation_MT}).
As we discussed in Section~\ref{sec:feature_analysis}, \textit{n\_words} and \textit{avg\_adj} are essential to \premodel accuracy,
it is clear that removing either of them severely deteriorates our \premodel.
If we were to keep \textit{avg\_sat\_adj}, we would see a 2.9\% increase in \premodel accuracy, however we choose to leave this out as it provides negligible improvements in the presence of BoW.

\cparagraph{Bag of words.} We found that including BoW as a feature in our \premodel brought improvements in accuracy for little overhead
(See Figure~\ref{fig:feature_analysis}b). We use the chi-squared test to evaluate each row of our BoW vector, and choose the top $k$
features. In Figure~\ref{fig:bow_feature_selection} we show the accuracy loss by choosing different values of $k$, as a baseline we use our
chosen value $k$=2000. It is clear that choosing a value greater than 2000 results in a dramatic loss in accuracy (nearly 4\%), which
quickly increases as $k$ increases. Setting $k$=1500 results in a small loss in accuracy, but reducing it further leads to much bigger
losses in accuracy. \eg $k$=500 results in a 5.75\% loss in accuracy. This indicates that $k$>2000 results in our \premodel that is prone
to overfitting, while $k$<1500 is unable to capture all of the information required for accurate predictions, therefore the optimal value
of $k$ sits around the 2000-1500 mark. We chose $k$=2000 as we achieved the highest accuracy with this value, and the overhead of
increasing $k$ is negligible.

\subsection{Training and Deployment Overhead \label{sec:overhead}}
Training the \premodel is a \emph{one-off} cost, and is dominated by the generation of training data which takes, in total, less than a day
(see Section~\ref{sec:premodel_training}). We can speed this up by using multiple machines. Compared to the training time of a typical
\DNN, our training overhead is negligible. Because our approach trades RAM space for improved accuracy and reduced inference time, we
provide an evaluation of resource utilization in Section~\ref{sec:resource_use}. In addition to our case studies, we have evaluated our
\premodel overhead for object detection, using the COCO dataset~\cite{coco}, where the runtime overhead is similar to image classification,
under 13.5\%.

\cparagraph{Image classification.} The runtime overhead of our \premodel is minimal, as depicted in Figure~\ref{fig:expermiental_results}a.
Out of a total average execution time of <1 second to classify an image, our \premodel accounts for 28\%. In comparison, this is 12.9\% and
71.7\% of the average execution time of the most (\mn{ResNet\_v1\_152}) and least (\mn{MobileNet\_v1\_100}) expensive models, respectively.
Furthermore, our energy footprint is smaller, making up 11\% of the total cost.
Comparing this to the most and least expensive models, gives an overhead of 7\% and 25\%, respectively.

\cparagraph{Machine translation.} Feature extraction costs are much smaller in this domain, hence the overheads of our \premodel are
negligible: <6ms overall, which accounts for 0.5\% of the end-to-end cost when translating a sentence. Similarly, the energy cost of our
\premodel accounts for 0.48\% of the overall energy cost. The memory footprint of our \premodel is also insignificant.

\begin{figure}[t!]
	\includegraphics[width=0.42\textwidth]{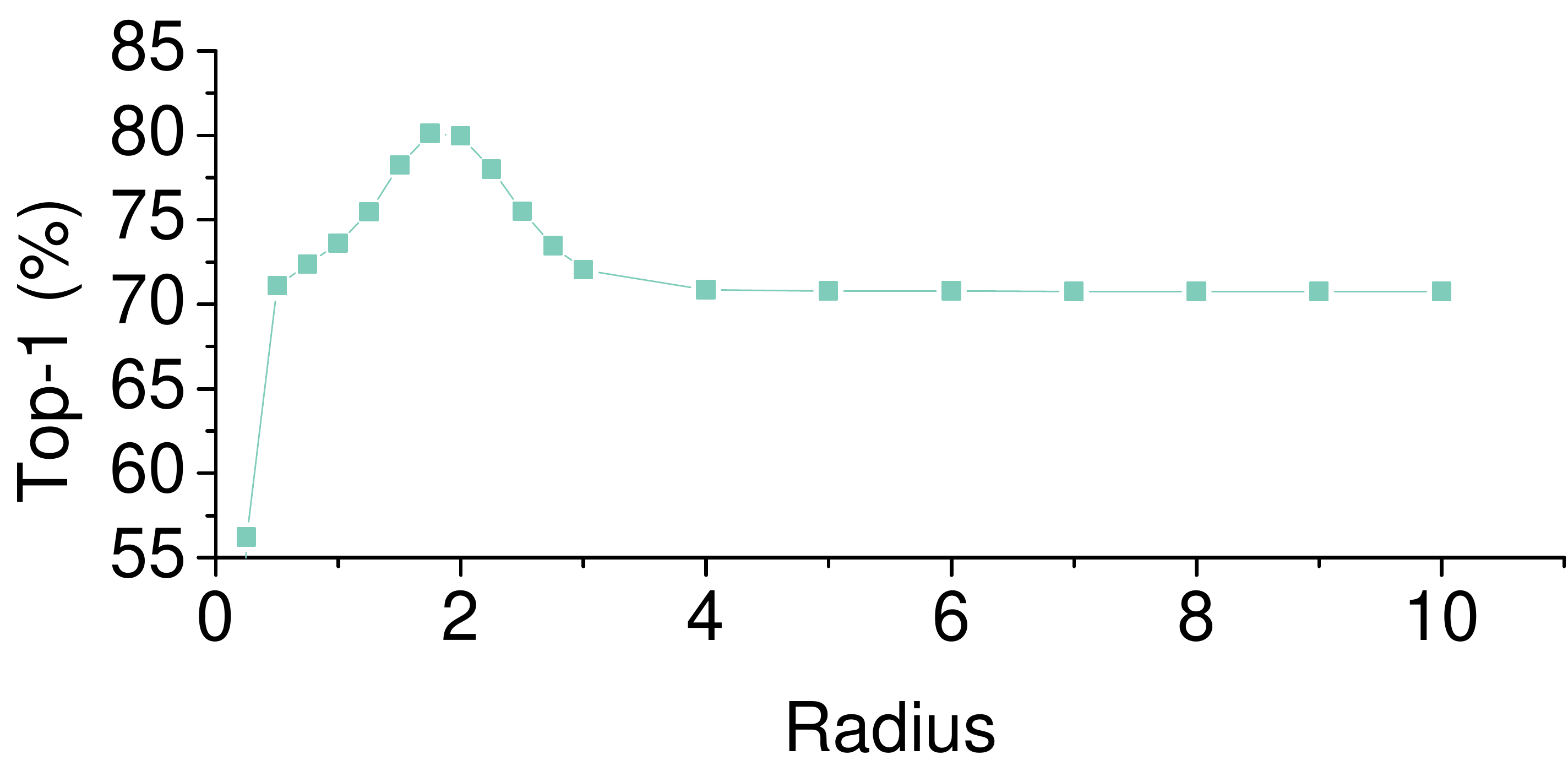}
	\vspace{-5mm}
	\caption{The \topone score when changing the radius of our image classification \premodel.}
    \vspace{-5mm}
	\label{fig:soundness_radius}
\end{figure}

\subsection{Soundness Analysis}
It is possible that our \premodel will provide an incorrect prediction. That is, it could choose either a \DNN that gives an incorrect
result, or a more expensive \DNN. Theoretical proof of soundness guarantee of machine learning models is an outstanding challenge and is
out of the scope of the paper~\cite{ bai2019rectified}. Nonetheless, there are two possible ways to empirically estimate the prediction
confidence: (1) using the distance on the feature space as a soundness measurement, or (2) using statistical assessments. We described both
methods as follows.

\cparagraph{Distance measurement.} Figure~\ref{fig:soundness_radius} shows how the accuracy of image classification (under the \topone
score) changes as the permissible distance for choosing the nearest training images changes. Recall that each training image is associated
with an optimal model for that image and by choosing the nearest training images to the input, we can then use a voting scheme to determine
which of the associate \DNNs to use for the input image. Here, the distance is calculated by computing the Euclidean distance between the
input testing image and a training image on the feature space. The results are averaged across our testing images using cross-validation.
When the permissible distance increases from 0 to 2, we see an increase in the inference accuracy. This is because using a short distance
reduces the chance of finding a testing image that is close enough. However, we observe that when the permissible distance is greater than
2, the inference accuracy drops as the distance increases. This is because when the permissible distance goes beyond this point,  we are
more likely to choose a testing image (and the associated optimal model) that is not similar enough to the input. This example shows that
the permissible distance can be empirically determined and used as a proxy for the accuracy confidence.

\cparagraph{Statistical assessments.} Another method for soundness guarantee is to combine probabilistic and statistical assessments. This
can be done by using a Conformal Predictor (CP)~\cite{shafer2008tutorial} to determine to what degree a \emph{new, unseen} input conforms
to previously seen training samples. The CP is a statistical assessment method for quantifying how much we could trust a model's
prediction. This is achieved by learning a nonconformity function from the model's training data. This function estimates the
``strangeness" from input features, $x$, to a prediction output, $y$, by looking at the input and the probability distribution of the model
prediction. Specifically, we learn a nonconformity function, $f$, from our \premodel training dataset, which produces a non-conformity
score for the \premodel's input $x_i$ and output $y_i$:
    $$
    \small
    f(x_i, y_i) = 1 - \hat{P}_h(y_i | x_i)
    $$

Here, $\hat{P}_h$ is the statistical distribution of the \premodel's probabilistic output, calculated as:
$$
\small
p_{x_i}^{y_i} = \frac{|\{z_j\in Z: a_j > a_{i}^{y_i}\}|} {q+1} + \theta \frac{|\{z_j\in Z: a_j = a_{i}^{y_i}\}| + 1}{q+1}, \theta\in[0,1]
$$
where $Z$ is part of the training dataset chosen by the CP, $q$ is the length of $Z$, $a_i$ is the calibration score learned from training
data, $a_{i}^{y_i}$ is the statistical score for \premodel prediction $y_i$, and $\theta$ is a calibration factor learned by the CP.

The learned function $f$ produces a non-conformity score between 0 and 1 for every class for each given input. The
closer the score to 0, the more likely the input is to conform to the \premodel's output, \ie it is similar to training
samples of that class. By choosing a threshold, we can predict whether our \premodel gives an incorrect \DNN for a
given input.  By implementing an SVM based conformal predictor for image classification, and using a threshold value of
0.5, we can correctly predict when our \premodel will choose an incorrect \DNN 87.4\% of the time, with a false
positive rate of 5.5\%. This experiment shows that we can use the CP to estimate if the \premodel's output can be
trusted to provides a certain degree of soundness guarantee.

\subsection{Further In-Depth Analysis}
This section contains an in-depth analysis using image classification as a case study. The results are similar when we apply the same
analysis to the machine translation case study.

\begin{figure*}[t!]
\centering
\begin{minipage}{.47\textwidth}
  \centering
  \includegraphics[width=.95\linewidth]{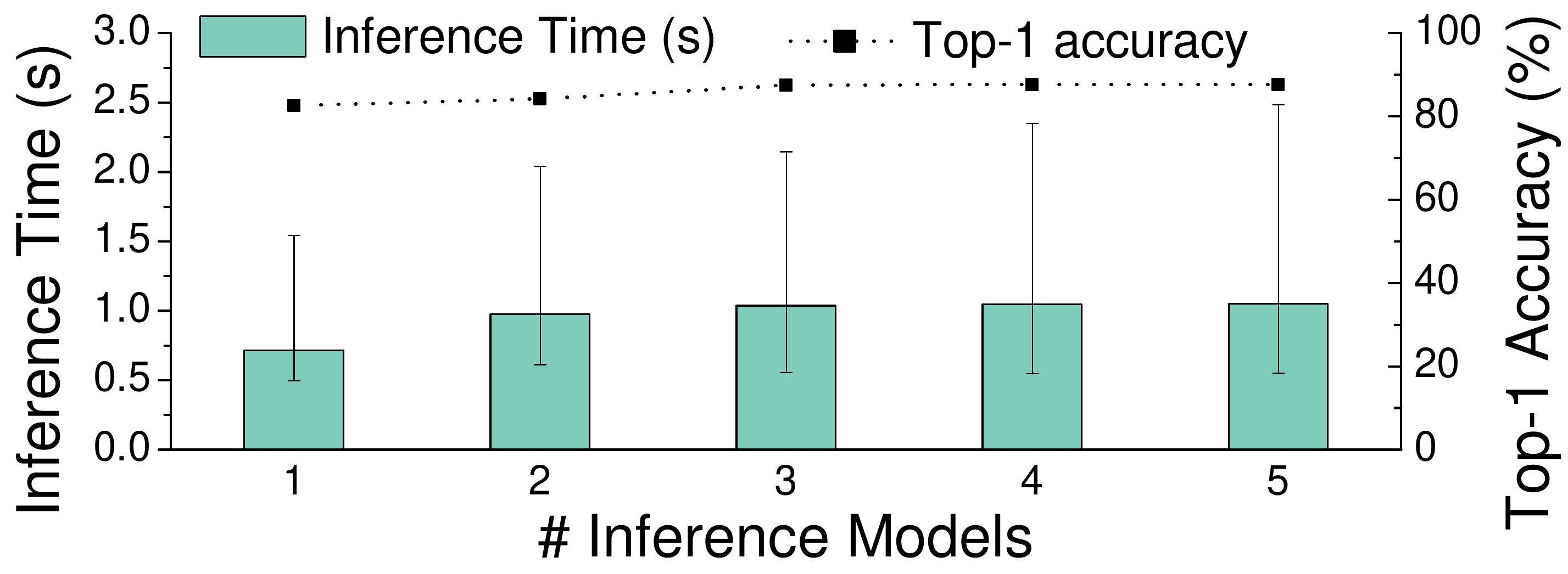}
  \caption{Overhead and achieved performance when using  a different number of \DNN models.
	The range of inference time across testing images is shown using min-max bars.}
	\label{fig:pre_model_different_levels}
\end{minipage}%
\hfill
\begin{minipage}{.47\textwidth}
  \centering
  \includegraphics[width=.95\linewidth]{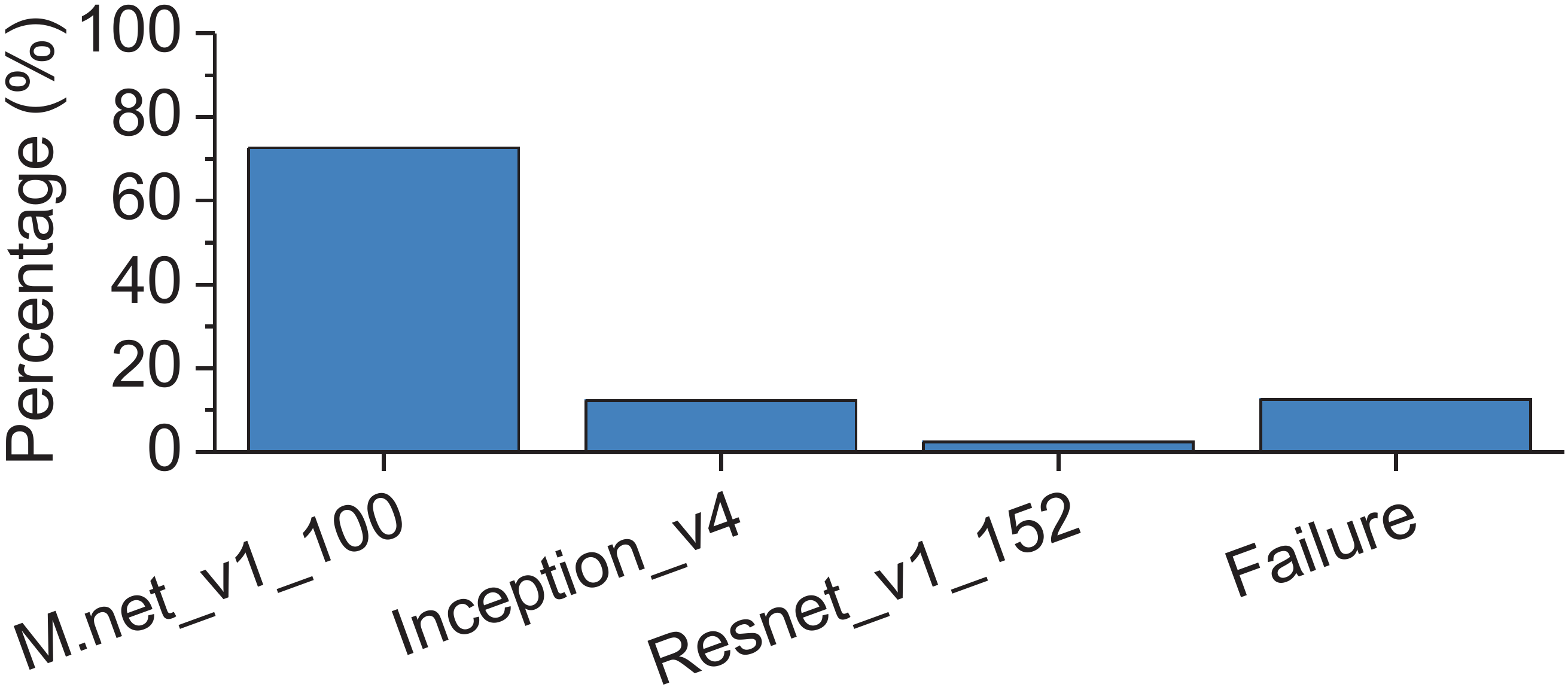}
  \caption{The utilization of each \DNN included in our \premodel.}
	\label{fig:pre_model_selection}
\end{minipage}

\begin{minipage}{1\textwidth}
		\centering
		\begin{tabularx}{1\textwidth} {>{\centering\arraybackslash}m{1.65in}>{\centering\arraybackslash}m{1.65in}>{\centering\arraybackslash}m{1.65in}}
			
			\includegraphics[width=0.310\textwidth,clip]{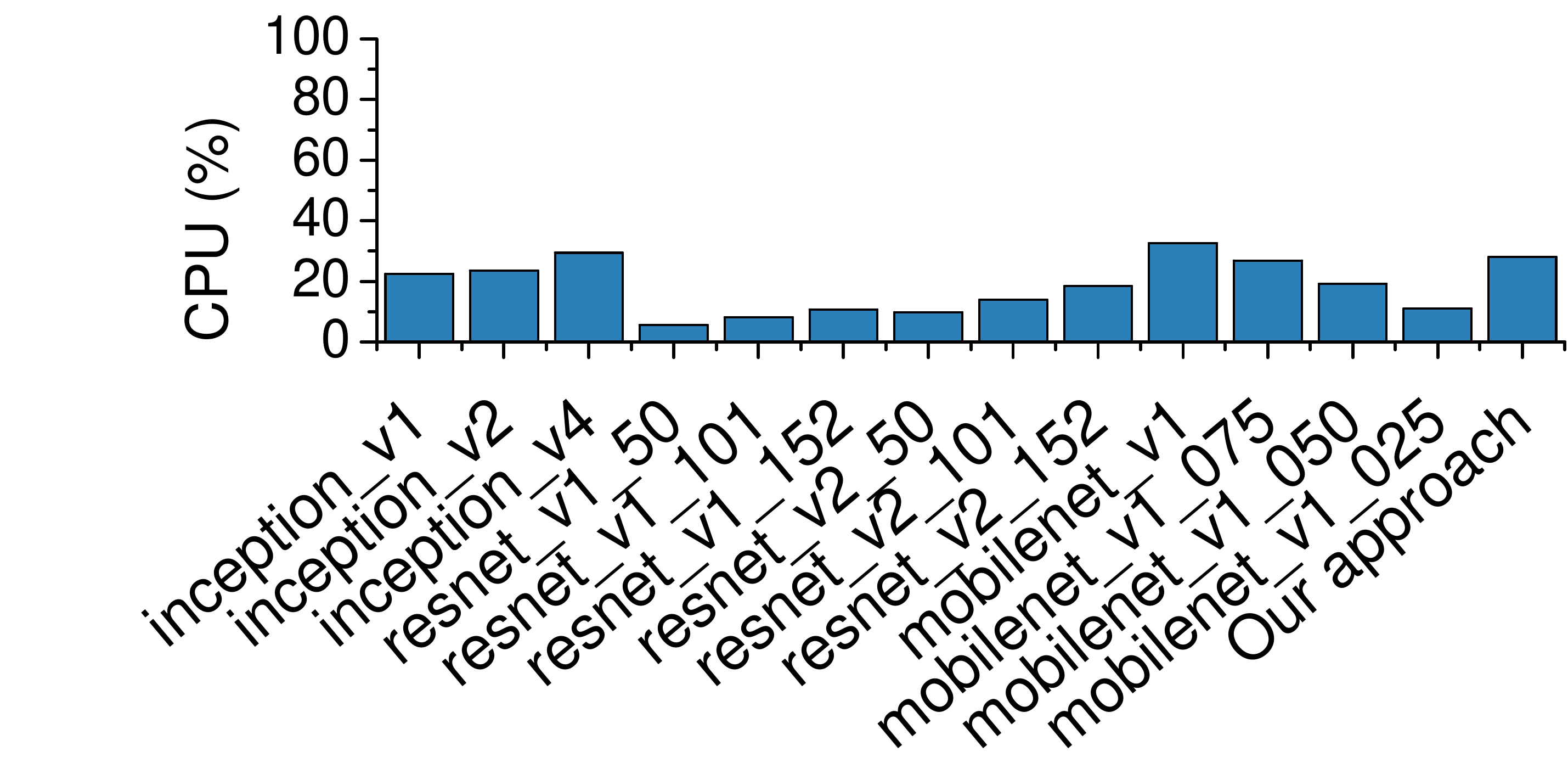} &
			\includegraphics[width=0.310\textwidth,clip]{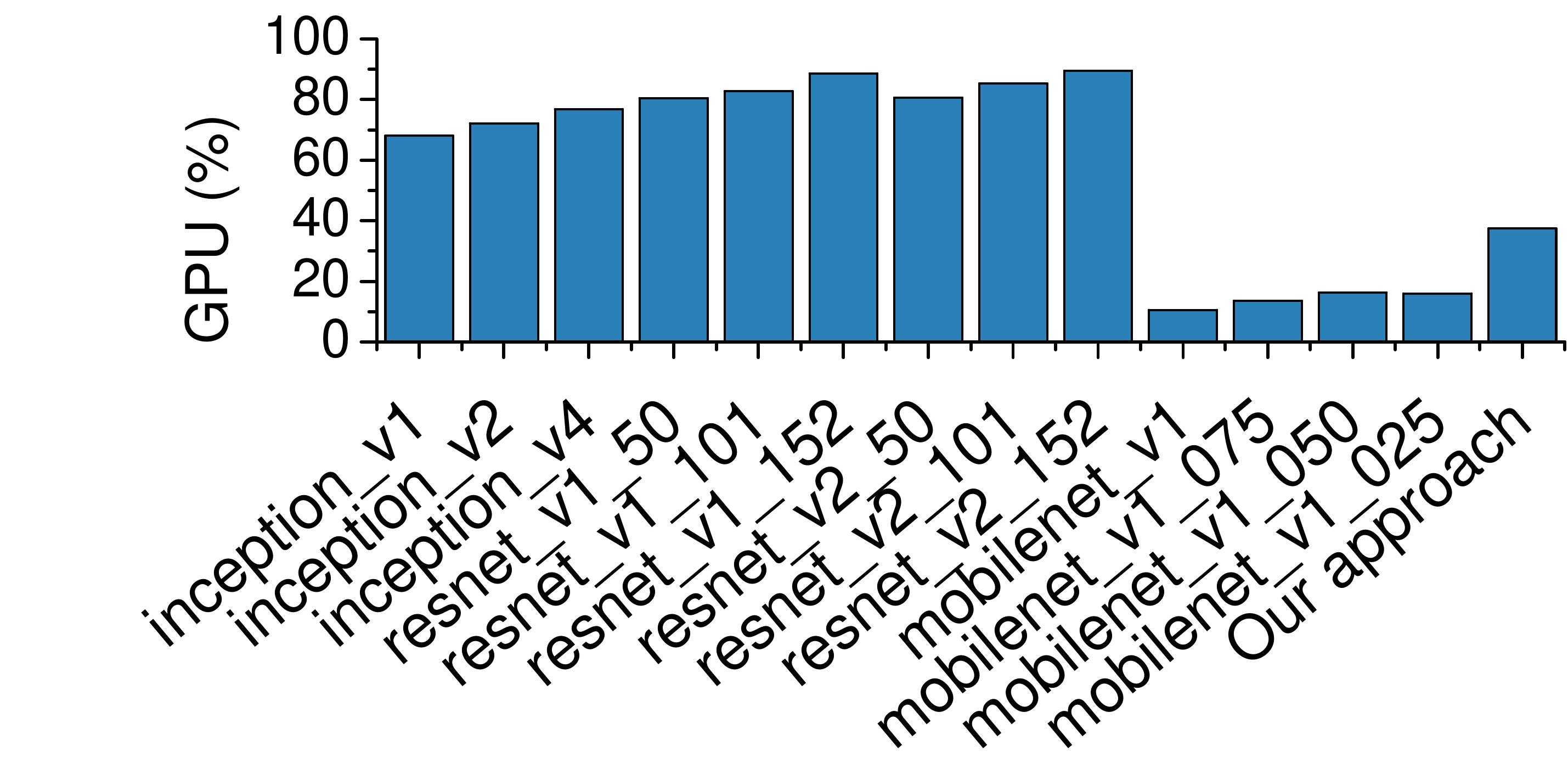} &	
			\includegraphics[width=0.310\textwidth,clip]{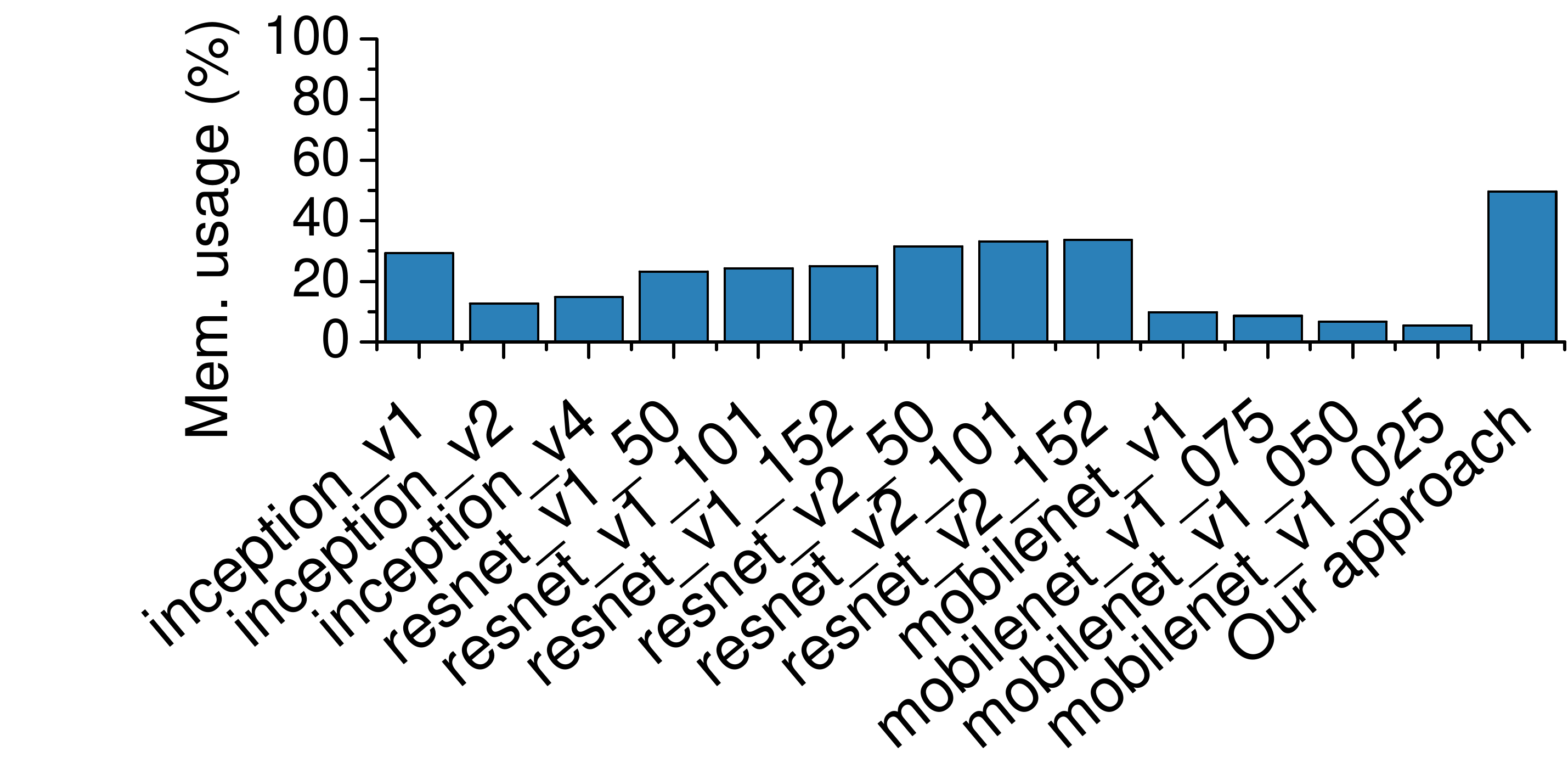} \\
			{\centering \scriptsize (a) CPU Utilization} &
			{\centering \scriptsize (b) GPU Utilization} &
			{\centering \scriptsize (c) Memory Utilization} 	\\
		\end{tabularx}
		\caption{The average CPU, GPU, and Memory utilization per model. Compared against our approach.}
		\label{fig:cpuGpuMemory}
\end{minipage}
\end{figure*}

\subsubsection{Changing the premodel Size}
In Section~\ref{sec:classifier_selection} we describe the method we use to chose which \DNN models to include. Using the \textit{Accuracy}
method, and temporarily ignoring the model selection threshold $\theta$ in Algorithm~\ref{alg:classifier_selection}, we constructed
Figure~\ref{fig:pre_model_different_levels}, where we compare the \topone accuracy and execution time using up to 5 \NN models. As we
increase the number of inference models, there is an increase in the end to end inference time as expensive models are more likely to be
chosen. At the same time, however, the \topone accuracy reaches a plateau of ($\approx$87.5\%) by using three \NN models. We conclude that
choosing three \NN models would be the optimal solution for our case, as we are no longer gaining accuracy to justify the increased cost.
This is in line with our choice of a value of 0.5 for $\theta$. Additionally, Figure~\ref{fig:pre_model_selection} shows the utilization
percentage of each model by our approach. Our approach can also choose to not select any model for an image if it deems none of the
available models as suitable for it. We use \emph{Failure} to represent this in the Figure. Overall, 87.5~\% of the time a model is
selected, leaving 12.5~\% of the time \textit{Failure} is selected.

In Section~\ref{sec:classifier_selection} we describe the method we use to chose which \DNN models to include.
Using the \textit{Accuracy} method, and
temporarily ignoring the model selection threshold $\theta$ in Algorithm~\ref{alg:classifier_selection}, we constructed Figure~\ref{fig:pre_model_different_levels},
where we compare the \topone accuracy and execution time using up to 5 \NN models.
As we increase the number of inference models, there is an increase in the end to end inference time as expensive models are more likely to be chosen.
At the same time, however, the \topone accuracy reaches a plateau of ($\approx$87.5\%) by using three \NN models.
We conclude that choosing three \NN models would be the optimal solution for our case, as we are no longer gaining accuracy to justify the increased cost.
This is in line with our choice of a value of 0.5 for $\theta$.
Additionally, Figure~\ref{fig:pre_model_selection} shows the utilization percentage of each model by our approach.
Our approach can also choose to not select any model for an image if it deems none of the available models as suitable for it.
We use \emph{Failure} to represent this in the Figure.
Overall, 87.5~\% of the time a model is selected, leaving 12.5~\% of the time \textit{Failure} is selected.

\vspace{-3mm}

%

\subsubsection{Resource Utilization} \label{sec:resource_use}
Figure~\ref{fig:cpuGpuMemory} shows the average CPU, GPU and memory utilization of a selection of image classification \DNNs. We recorded
the utilization of each resource during inference on every image in the ImageNet ILSVRC 2012 \emph{validation dataset}, and report the
average.

\cparagraph{CPU.} Figure~\ref{fig:cpuGpuMemory}a shows the CPU utilization. All \DNNs primarily run on the GPU, therefore we see a low CPU
utilization overall; no \DNN has a utilization higher than 30\%. Our approach is one of the most expensive, using 28.11\% of the CPU, it is
only cheaper than \mn{MobileNet\_v1} and \mn{Inception\_v4}, which use 32.63\% and 29.42\%, respectively. In this category, our approach is
expensive as we include the two most expensive models.

\cparagraph{GPU.} GPU utilization is shown in Figure~\ref{fig:cpuGpuMemory}b. As expected, this is much higher than CPU utilization, with
the majority of \DNNs using between 70-90\% of the GPU. In contrast, our approach has a much lower utilization of 37.46\%; 52.18\% lower
than the most expensive model, \mn{ResNet\_v2\_152}. We achieve this by making use of \mn{MobileNet\_v1} whenever possible, which has a
utilization of 10.57\%.

\cparagraph{Memory.} Figure~\ref{fig:cpuGpuMemory}c compares the memory utilization. Our approach keeps the selected \DNNs in memory,
therefore it is the most expensive in this category. However, our approach only requires 16\% more memory than the most expensive model, a
small cost to pay for reduced CPU and GPU load, and a faster inference time with higher accuracy.

%
%
\vspace{-3mm}
\subsubsection{Compression} \label{sec:compression_analysis}
So far we have shown the ability of our approach to utilize multiple \DNNs, however, this is not always possible.
In some cases only a single trained \DNN is available, this could be caused by a number of reasons, \eg limited training time.
This section shows how our approach can still be utilized in this case, by making use of compression.
We use two different compression algorithms:  Deep Compression~\cite{hansong2015} and Quantization~\cite{jacob2018quantization}.
By first applying Deep Compression followed by Quantization, we effectively have a third compression "algorithm".
Compression is designed to make a \DNN \emph{lighter} -- it has a faster inference time and a smaller size (See Table~\ref{tbl:compression_info}) -- however, as a consequence the model accuracy also degrades.

We chose $Resnet\_v2\_152$ as a starting model.
It is the most complex model we consider with the highest accuracy, unfortunately, as a consequence it also has the longest runtime at 2026ms.
By applying each of our three compression algorithms, we generate a total of 4 \emph{different} models.
Using the 4 distinct \DNNs, we apply our method to create a new \premodel.

\begin{table}
	\begin{minipage}{0.45\linewidth}
		\caption{The change in model size when using compression on $Resnet\_v2\_152$}
		\label{tbl:compression_info}
\vspace{-2mm}
		\centering
        \scriptsize
		\begin{tabular}{ll}
			\toprule
			\textbf{Model}                  & \textbf{Size (MB)}         \\
			\midrule
			\rowcolor{Gray}     Without Compression   & 691                        \\
			Deep Compression      & 317.12                     \\
			\rowcolor{Gray}     Quantization          & 473.42                     \\
			Both Compression Methods     & 226.22                     \\
			\bottomrule
		\end{tabular}
	\end{minipage}\hfill
	\begin{minipage}{0.45\linewidth}
		\centering
		\includegraphics[width=\textwidth]{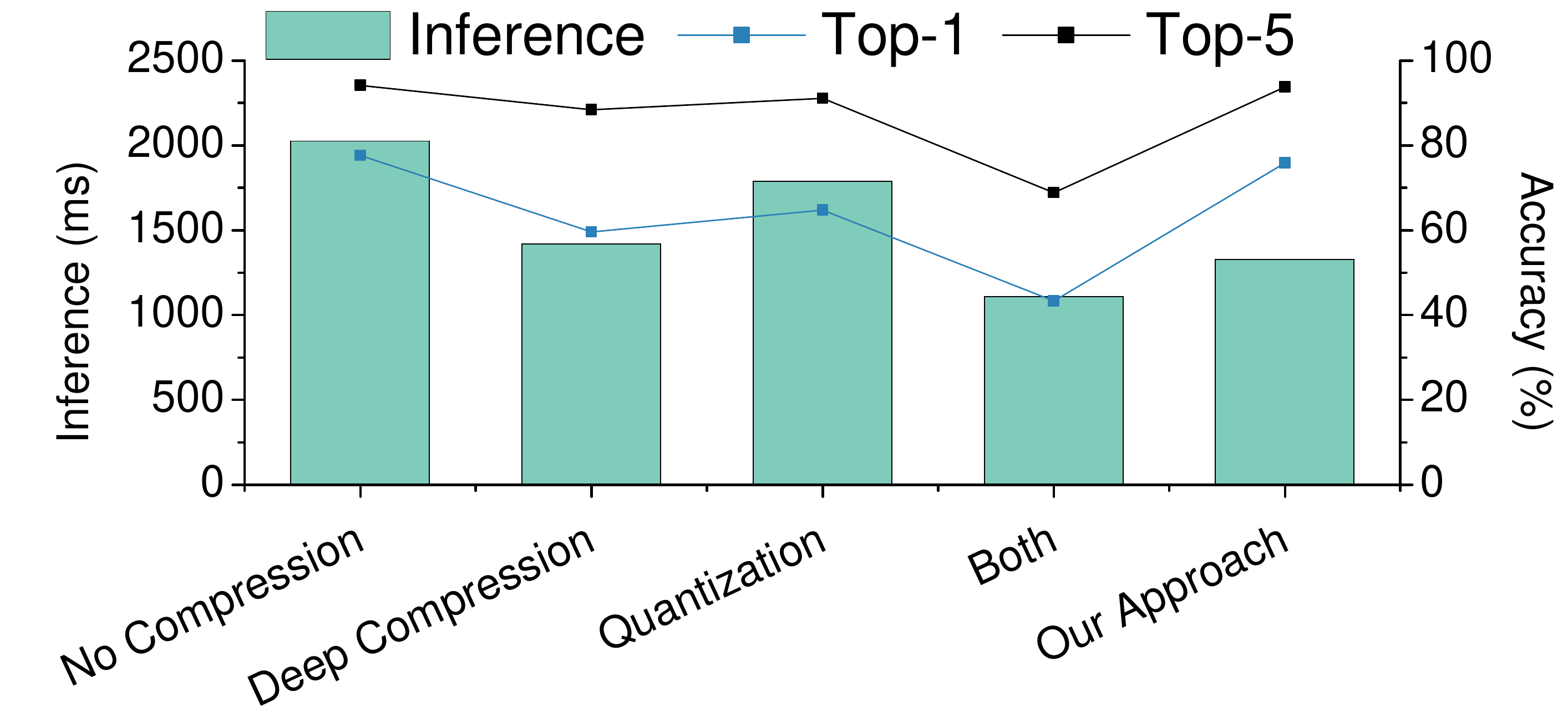}\\
		\vspace{-5mm}
		\captionof{figure}{The inference time, \topone, and \topfive performance using compression on a single \DNN.}
		\label{fig:compression_comp}
	\end{minipage}
	\vspace{-4mm}
\end{table}

Figure~\ref{fig:compression_comp} shows the performance of each compressed model and our approach. There is a clear trend, applying
compression reduces accuracy while reducing inference time. Applying both compression methods in practice would result in an unacceptable
accuracy drop, reducing \topone accuracy by 34.32\%. However, it makes sense in this scenario as our approach is able to make use of a
model compressed by both methods when it can meet the accuracy constraint. Our approach is able to achieve a minor drop in accuracy (1.76\%
for \topone, and 0.31\% for \topfive), while reducing inference time by 1.52x. Effectively, we are able to utilize the positive of
compression (reduced runtime) while keeping the accuracy of the original model.

\section{Discussion} \label{sec:discussion}
Naturally, there is  room for further work and possible improvements. We discuss a few points here.

 \cparagraph{Feature extraction.} The
majority of our image classification overhead is caused by feature extraction
 for our \premodel.
 Our prototype feature extractor is written in Python; by re-writing this tool in a more
efficient language can reduce the overhead.
There are also hotshots in our code which would benefit from parallelism.

\cparagraph{Premodel training.} There is room for improvement for our machine translation \premodel. We were unable to reach the full
potential shown by the \oracle. To aid the \premodel in reaching its full potential would require improving its accuracy. We believe we
require more training data. There was only 5K sentences available for machine translation, in comparison to 50k images for image
classification.

\cparagraph{Computation Offloading.} This work focuses on accelerating inference on the current device. Future work could involve an
environment with the opportunity to offload some of the computation to either cloud servers, or other devices at the
edge~\cite{elkhatib2017microclouds}. Accomplishing this would require a method to measure and predict network latency, allowing an educated
decision to be made at runtime. ML techniques are shown to be effective in learning a cost function for profitability
analysis~\cite{grewe2013portable}. This can be integrated with our current learning framework.

\cparagraph{Processor choice.} By default, inference is carried out on a GPU, but this may not always be the best choice.  Previous work
has already shown ML techniques to be successful at selecting the optimal computing device~\cite{taylor2017adaptive}. This can be
integrated into our existing learning framework.

\cparagraph{Model size.} Our approach uses multiple pre-trained \DNN models for inference, in comparison, the default method is to  simply
use a single model. Therefore, our approach requires more storage space. A solution for this would involve using model compression
techniques to generate multiple compressed models from a single, highly accurate model. We have shown that our approach is effective at
choosing between compressed models. The result of this is numerous models sharing many weights in common, which allows us to amortize the
cost of using multiple models.


\section{Related Work} \label{sec:related_work}


Methods have been proposed to reduce the computational demands of a deep model by trading prediction accuracy for
runtime, compressing a pre-trained
network~\cite{han2015learning,Chen:2015:CNN:3045118.3045361,DBLP:journals/corr/RastegariORF16}, training small networks
directly~\cite{projectnet,Georgiev:2017:LMA:3139486.3131895}, or a combination of both~\cite{howard2017mobilenets}.
Using these approaches, a user now needs to decide when to use a specific model.
Making such a crucial decision is a non-trivial task as the application context (\eg the model input) is often unpredictable and
constantly evolving. Our work alleviates this user burden by automatically selecting an appropriate model to use.

Neurosurgeon \cite{Kang2017neurosurgeon} identifies when it is beneficial (\eg in terms of energy consumption and end-to-end latency) to offload a \DNN layer to be computed on the cloud. Unlike
Neurosurgeon, we aim to minimize \emph{on-device} inference time without compromising prediction accuracy.
The work presented by \citet{RodriguezWZMH17} trains a model twice; once on shared data and again on personal data, in an attempt to
prevent personal data being sent outside the personal domain. In contrast to the latter two works, our approach allows having a diverse set
of networks, by choosing the most effective network to use at runtime. They, however, are complementary to our approach, by providing the
capability to fine-tune a single network structure.

Recently, a number of software-based approaches have been proposed to accelerate \CNNs on embedded devices. They aim to accelerate
inference time by exploiting parameter tuning~\cite{latifi2016cnndroid}, computational kernel optimization~\cite{han2016eie,
bhattacharya2016sparsification}, task parallelism~\cite{Motamedi:2017:MIR:3145508.3126555,rallapalli2016very}, and trading precision for
time~\cite{Huynh:2017:DMG:3081333.3081360} etc. Since a single model is unlikely to meet all the constraints of accuracy, inference time
and energy consumption across inputs~\cite{guo2017towards}, it is attractive to have a strategy to dynamically select the appropriate model
to use. Our work provides exactly such a capability and is thus complementary to these prior approaches.

Off-loading computation to the cloud can accelerate \DNN model inference \cite{teerapittayanon2017distributed}, but this is not always
applicable due to privacy, latency or connectivity issues. The work presented by Ossia \etal partially addresses the issue of
privacy-preserving when offloading \DNN inference to the cloud ~\cite{ossia2017hybrid}. Our adaptive model selection approach allows one to
select which model to use based on the input, and is also useful when cloud offloading is prohibitively.

Machine learning has been employed for various optimization tasks, including code
optimization~\cite{cummins2017end,mlcpieee,wang2014integrating,Tournavitis:2009:THA:1542476.1542496,Wang:2009:MPM:1504176.1504189,wang2010partitioning,grewe2013portable,wang2013using,DBLP:journals/taco/WangGO14,taylor2017adaptive,
ogilvie2014fast,cbenchmarks,ogilvie2017minimizing,ipdpsz18,spmv,ijpp19}, task
scheduling~\cite{grewe2011workload,emani2013smart,grewe2013opencl,Delimitrou:2014:QRQ:2541940.2541941,marco2017improving,ren2017optimise,conext18,yuan2019using},
cloud deployment~\cite{Samreen2016Daleel,Samreen2019tl}, network management~\cite{abs-1709-06599}, \etc. Our approach is closely related to
ensemble learning where multiple models are used to solve an optimization problem. This technique is shown to be useful on scheduling
parallel tasks~\cite{Emani:2015:CDM:2737924.2737999}, wireless sensing~\cite{mobicom18}, and optimize application memory
usage~\cite{Marco:2017:ISA:3135974.3135984}. This work is the first attempt in applying this technique to optimize deep inference on
embedded devices.

Many significantly notable improvements have been made for machine translation over the last few years, including
Google Neural Machine Translation~\cite{45610}, and the introduction of the Attention
architecture~\cite{vaswani2017attention}. A common method to improve machine translation accuracy is
ensembling~\cite{stahlberg2018university,saunders2018multi}, where multiple models are used during one translation. Our
approach is able to see improvements in accuracy without the added cost of ensembling; we only run one translator model
for each translation task. In recent years \CNNs have become the norm for sentence classification.
\cite{kim2014convolutional} shows that even simple \CNNs can be used classify sentences with high accuracy, however
running a \CNN on embedded systems is expensive. \citet{joulin2017bag} explore a simple, fast text classifier.
Unfortunately, this classifier leads to poor performance on our data.

\section{Conclusion}
We have presented a novel approach for efficient deep learning inference for embedded systems. Our approach leverages multiple \DNNs
through the use of a \premodel that dynamically selects the optimal model to use, depending on the model input and evaluation criterion. We
developed an automatic approach for \premodel generation as well as feature selection and tuning. We apply our approach to two deep
learning domains: image classification and machine translation, which involve convolutional and recurrent neural network architectures.
Experiment results show that our approach deliver portable good performance across application domains and neural network architectures.
For image classification, our approach achieves an overall \topone accuracy of above 87.44\%, which translates into an improvement of
7.52\% and 1.8x reduction in inference time when compared to the most-accurate single deep learning model. For machine translation, our
approach is able to reduce inference time by 1.34x than the single most capable model, without significantly effecting accuracy. With more
training data we could achieve the same reduction in accuracy while increasing F1 measure by 20.51\%.




\section*{Acknowledgement}
This work was partly supported by the UK EPSRC under grants EP/M015734/1 (Dionasys) and EP/M01567X/1 (SANDeRs). For any correspondence,
please contact Zheng Wang (E-mail: z.wang5@leeds.ac.uk).

{\small
\bibliographystyle{ACM-Reference-Format}
\bibliography{bibs,zheng}
}

\end{document}